\documentclass[10pt,twocolumn,letterpaper,table,dvipsnames]{article}
\usepackage[accsupp]{axessibility}
\usepackage{wacv}
\usepackage{times}
\usepackage{epsfig}
\usepackage{amsmath}
\usepackage{amssymb}
\usepackage[ruled,vlined]{algorithm2e}

\usepackage{soul}
\usepackage{graphicx}

\usepackage{bm}
\usepackage{xcolor}
\usepackage{stmaryrd}
\makeatletter
\@namedef{ver@everyshi.sty}{}
\makeatother
\usepackage{subcaption}
\usepackage{mathtools}

\newcommand{\rb}{\rrbracket}
\newcommand{\lb}{\llbracket}

\usepackage{booktabs}
\usepackage{multirow}

\usepackage{xspace} 
\def\bic{\textit{BiC}\xspace}
\def\adbic{\textit{adBiC}\xspace}
\def\lwf{\textit{LwF}\xspace}
\def\lucir{\textit{LUCIR}\xspace}
\def\siw{\textit{SIW}\xspace}
\def\ftplus{\textit{FT}+\xspace}

\def\imnet{\textsc{Imn}-100\xspace}
\def\cifar{\textsc{Cifar}-100\xspace}
\def\birds{\textsc{Birds}-100\xspace}
\def\food{\textsc{Food}-100\xspace}
\def\places{\textsc{Places}-100\xspace}

\definecolor{dgreen}{rgb}{0.05,0.66,0.}

\newcommand{\tabnoteb}[1]{{\footnotesize #1}}
\newcommand{\tabnoteg}[1]{{\color{dgreen} \footnotesize \textbf{#1}}}
\newcommand{\tabnoter}[1]{{\color{red} \footnotesize \textbf{#1}}}
\newcommand\cmdots{\hbox to 1em{.\hss.\hss.}}

\usepackage[ruled,vlined]{algorithm2e}
\usepackage{algpseudocode}
\usepackage{amsmath}
\algnewcommand\algorithmicforeach{\textbf{for each}}
\algdef{S}[FOR]{ForEach}[1]{\algorithmicforeach\ #1\ \algorithmicdo}

\usepackage{graphicx}
\newcommand\sbullet[1][.5]{\mathbin{\vcenter{\hbox{\scalebox{#1}{$\bullet$}}}}}

\usepackage{multicol} 

\usepackage{upgreek}

\usepackage{tabularx}

\def\wacvPaperID{206}

\wacvfinalcopy

\usepackage[breaklinks=true,bookmarks=false,colorlinks]{hyperref}
\hypersetup{citecolor=green}
\begin{document}


\title{Dataset Knowledge Transfer for Class-Incremental Learning without Memory}

\author{Habib Slim$^{1}$\thanks{Equal contribution}\hspace{1em}
Eden Belouadah$^{1, 2}$\footnotemark[1]\hspace{1em}
Adrian Popescu$^{1}$\hspace{1em}
Darian Onchis$^{3}$ \\
$^{1}$ Université Paris-Saclay, CEA, List, F-91120, Palaiseau, France\\
$^{2}$ IMT Atlantique, Lab-STICC, team RAMBO, UMR CNRS 6285, F-29328, Brest, France\\
$^{3}$ West University of Timisoara, Timisoara, Romania \\
{\tt\footnotesize habib.slim@grenoble-inp.org, \{eden.belouadah, adrian.popescu\}@cea.fr, darian.onchis@e-uvt.ro}
}

\maketitle
\thispagestyle{empty}

\begin{abstract}

Incremental learning enables artificial agents to learn from sequential data. 
While important progress was made by exploiting deep neural networks, incremental learning remains very challenging.
This is particularly the case when no memory of past data is allowed and catastrophic forgetting has a strong negative effect. 
We tackle class-incremental learning without memory by adapting prediction bias correction, a method which makes  predictions of past and new classes more comparable.
It was proposed when a memory is allowed and cannot be directly used without memory, since samples of past classes are required.
We introduce a two-step learning process which allows the transfer of bias correction parameters between reference and target datasets.
Bias correction is first optimized offline on reference datasets which have an associated validation memory.
The obtained correction parameters are then transferred to target datasets, for which no memory is available.
The second contribution is to introduce a finer modeling of bias correction by learning its parameters per incremental state instead of the usual past vs. new class modeling.
The proposed dataset knowledge transfer is applicable to any incremental method which works without memory. 
We test its effectiveness by applying it to four existing methods.
Evaluation with four target datasets and different configurations shows consistent improvement, with practically no computational and memory overhead. 

\end{abstract}


\vspace{-5mm}

\section{Introduction}
\label{sec:intro}

\begin{figure*}[t]
	\begin{center}
    \includegraphics[width=0.8\textwidth]{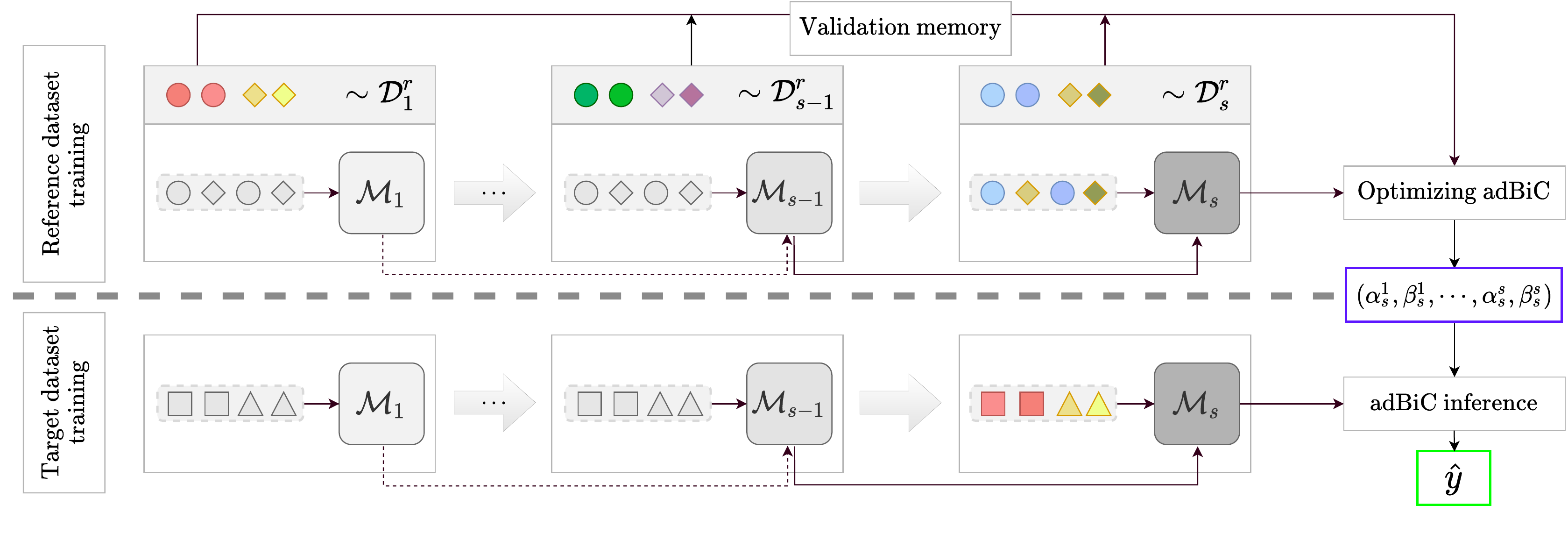}
    \vspace{-0.8cm}
	\end{center}
	\caption{Illustration of \textit{TransIL}, our proposed method, depicting states from $1$ to $s$ for a reference and a target dataset. The model $\mathcal{M}$ is updated in each state with data from new classes. States from $1$ to $s-1$ are faded to convey the fact that knowledge learned in them is affected by catastrophic forgetting.
	The class IL process is first launched offline on the reference dataset where \adbic, our proposed bias correction layer, is trained using a validation memory which stores samples for past and new classes. 
	Class IL is then applied to the target dataset, but without class samples shared across states since a memory is not allowed in this scenario.
	The set of optimal parameters of \adbic obtained for the reference dataset is transferred to the target dataset.
	This is the only information shared between the two processes and it has a negligible memory footprint. 
	The transfer of parameters enables the use of bias correction for the target dataset.
	The final predictions obtained in state $s$ are improved compared to the direct use of $\mathcal{M}_s$ predictions, since the bias in favor of new classes is reduced. 
	}
	\vspace{-4mm}
	\label{fig:overview}
\end{figure*}

Incremental learning (IL) enables the adaptation of artificial agents to dynamic environments in which data is presented in streams.
This type of learning is needed when access to past data is limited or impossible, but is affected by catastrophic forgetting~\cite{mccloskey1989_catastrophic}.
This phenomenon consists in a drastic performance drop for previously learned information when ingesting new data. 
Works such as~\cite{castro2018_e2eil,douillard2020_podnet,hou2019_lucir,rebuffi2017_icarl,wu2019_bic,zhao2020_maintaining,zhou2019_m2kd} alleviate the effect of forgetting by replaying past data samples when updating deep incremental models in class IL. 
A term which adapts knowledge distillation~\cite{hinton2015_distillation} to IL is usually exploited to reinforce the representation of past classes~\cite{li2016_lwf}.
When such a memory is allowed, class IL actually becomes an instance of imbalanced learning~\cite{he2009_imbalanced}.
New classes are favored since they are represented by a larger number of images.
As a result, classification bias correction methods were successfully introduced in~\cite{castro2018_e2eil,wu2019_bic,zhao2020_maintaining}. 

While important progress was made when a fixed memory is allowed, this is less the case for class IL without memory.
This last setting is more challenging and generic since no storage of past samples is allowed.
In absence of memory, existing methods become variants of \textit{Learning without Forgetting} ($LwF$)~\cite{li2016_lwf} with different formulations of the distillation term. 
Importantly, bias correction methods become inapplicable without access to past classes samples. 

Our main contribution is to enable the use of the bias correction methods, such as the \bic layer from~\cite{wu2019_bic}, in class IL without memory.
We focus on this approach because it is both simple and effective in IL with memory~\cite{belouadah2021_study, masana2021_study}. Authors of \bic~\cite{wu2019_bic} use a validation set which stores samples of past classes to optimize parameters. 
Instead, we learn correction parameters offline on a set of reference datasets and then transfer them to target datasets.
The method is thus abbreviated \textit{TransIL}.
The intuition is that, while datasets are different, optimal bias correction parameters are stable enough to be transferable between them. 
We illustrate the approach in Figure~\ref{fig:overview}, 
with the upper showing the IL process with a reference dataset.
A memory for the validation samples needed to optimize the bias correction layer is allowed since the training is done offline.
The lower part of the figure presents the incremental training of a target dataset.
The main difference with the standard memoryless IL training comes from the use of a bias correction layer optimized on the reference dataset.
Its introduction leads to an improved comparability of prediction scores for past and new classes.
Note that the proposed method is applicable to any class IL method, since it only requires the availability of raw predictions provided by deep models $\mathcal{M}_s$. 

The second contribution is to refine the definition of the bias correction layer introduced in~\cite{wu2019_bic}. 
The original formulation considers all past classes equally in the correction process.
With~\cite{masana2021_study}, we hypothesize that the degree of forgetting associated to past classes depends on the initial state in which they were learned. 
Consequently, we propose \textit{Adaptive BiC} (\adbic), an optimization procedure which learns a pair of parameters per IL state instead of a single pair of parameters as proposed in~\cite{wu2019_bic}.

We provide a comprehensive evaluation of \textit{TransIL} by applying it to four backbone class IL methods. 
Four target datasets with variable domain shift with respect to reference datasets and different numbers of IL states are used.
An improvement of accuracy is obtained for almost all tested configurations. 
The additional memory needs are negligible since only a compact set of correction parameters is stored.
Code and data needed for reproducibility are provided\footnote{\url{https://github.com/HabibSlim/DKT-for-CIL}}.
\section{Related work}
\label{sec:sota}
Incremental learning is a longstanding machine learning task~\cite{fritzke1994_growing, martinetz_1993, syed1999_handling} which witnessed a strong growth in interest after the introduction of deep neural networks.
It is named differently as continual, incremental or lifelong learning depending on the research communities which tackle it and the setting of the problem. 
However, the objective is common: enable artificial agents to learn from data which is fed sequentially to them. 
Detailed reviews of existing approaches are proposed, among others, in~\cite{belouadah2021_study,lange_2019, masana2021_study, parisi2019_continual}. 
Here, we analyze works most related to our proposal, which tackle class IL and keeps memory and computational requirements constant, or nearly so, during the IL process.
We focus particularly on methods which address bias in favor of new classes~\cite{masana2021_study} and were designed for class IL with memory.

The wide majority of class IL methods make use of an information preserving penalty \cite{dhar2018_lwm}.
This penalty is generally implemented as a loss function which reduces the divergence between the current model and the one learned in the preceding IL state.
Learning without forgetting (\lwf)~\cite{li2016_lwf} is an early work which tackles catastrophic forgetting in deep neural nets.
It exploits knowledge distillation~\cite{hinton2015_distillation} to preserve information related to past classes during incremental model updates.
Less-forgetting learning~\cite{jung2016_less} is a closely related method.
Past knowledge is preserved by freezing the softmax layer of the source model and updating the model using a loss which preserves the representation of past data.
The two methods aim to propose a good compromise between plasticity, needed for new data representation, and stability, useful for past information preservation.
However, they require the storage of the preceding model in order to perform distillation toward the model which is currently learned. 
This requirement can be problematic if the memory of artificial agents is constrained. 

\lwf was initially used for task-based continual learning and was then widely adopted as backbone for class IL. 
\textit{iCaRL}~\cite{rebuffi2017_icarl} exploits \lwf and a fixed-size memory of past samples to alleviate catastrophic forgetting.
In addition, a nearest-mean-of-exemplars classifier is introduced in order to reduce the bias in favor of new classes.
\textit{E2EIL}~\cite{castro2018_e2eil} corrects bias by adding a second fine-tuning step with the same number of samples for each past and new class.
The learning of a unified classifier for incremental learning rebalancing (\lucir) is proposed in~\cite{hou2019_lucir}.
The authors introduce a cosine normalization layer in order to make the magnitudes of past and new class predictions more comparable.
The maintenance of both discrimination and fairness is addressed in~\cite{zhao2020_maintaining}. 
The ratio between the mean norm of past and new class weights is applied to the weights of new classes, to make their associated predictions more balanced. 
Bias Correction (\bic) \cite{wu2019_bic} exploits a supplementary linear layer to rebalance predictions of a deep incremental model.
A validation set is used to optimize the parameters of this linear layer, which modifies the predictions of the deep model learned in a given incremental state. 
We tackle two important limitations of existing bias correction methods.
First, they are inapplicable without memory because they require the presence of past class samples. 
We propose to transfer bias correction layer parameters between datasets to address this problem.
Second, the degree of forgetting associated to past classes is considered equivalent, irrespective of the initial state in which they were learned.   
This is problematic insofar as a recency bias, which favors classes more recently, appears in class IL~\cite{masana2021_study}.
We refine the linear layer from~\cite{wu2019_bic} to improve the handling of recency bias. 

The improvement of the component which handles model stability also received strong attention in class IL.
Learning without memorizing~\cite{dhar2018_lwm} is inspired by \lwf and adds an attention mechanism to the distillation loss.
This new term improves the preservation of information related to base classes.
A distillation component which exploits information from all past states and from intermediate layers of CNN models was introduced in~\cite{zhou2019_m2kd}.
\lucir~\cite{hou2019_lucir} distills knowledge in the embedding space rather than the prediction space to reduce forgetting and adds an inter-class separation component to better distinguish between past and new class embeddings. 
\textit{PODNet}~\cite{douillard2020_podnet} employs a spatial-based distillation loss and a representation which includes multiple proxy vectors for classes to optimize distillation.
In~\cite{kumar2021_efficient}, a feature map transformation strategy with additional network parameters is proposed to improve class separability. 
Model parameters are shared between global and task-specific parameters and only the latter are updated at each IL state to improve training times.
Feature transformation using a dedicated MLP is introduced in~\cite{iscen2020_memory}. 
This approach only stores features but adds significant memory to store the additional MLP.
Recently, the authors of~\cite{kurmi2021_not} argued for the importance of uncertainty and of attention mechanisms in the modeling of past information in class IL.
These different works provide a performance gain compared to the original adaptation of distillation for continual learning~\cite{li2016_lwf} in class IL with memory.

The utility of distillation in a class IL scenario was recently questioned.
It is shown~\cite{masana2021_study,prabhu2020gdumb} that competitive results are obtained if a fixed-size memory is allowed for large-scale datasets. 
The distillation component is removed in~\cite{masana2021_study} and IL models are updated using fine-tuning.
A simpler approach is tested in~\cite{prabhu2020gdumb}, where the authors learn models independently for each incremental state after balancing class samples. 
The usefulness of distillation was also challenged in absence of a memory~\cite{belouadah2020_siw} where  standardization of initial weights (\siw), learned when a class was first encountered, was proposed in~\cite{belouadah2020_siw}. The freezing of initial weights was tested in~\cite{masana2021_study} and also provides significant improvements.
It is thus interesting to also apply the proposed approach to methods which do not exploit distillation.

Our method is globally inspired by existing works which transfer knowledge between datasets. 
We mentioned knowledge distillation~\cite{hinton2015_distillation} which is widely used in IL.
Dataset distillation~\cite{wang2018_dataset} encodes large datasets into a small set of synthetic data points to make the training process more efficient.
Hindsight anchor learning~\cite{chaudhry2020_using} learns an anchor per class to characterize points which would maximize forgetting in later IL states.
While the global objective is similar, our focus is different since only a very small number of parameters are transferred from reference to target datasets to limit catastrophic forgetting on the latter.

\section{Dataset knowledge transfer for class IL}
In this section, we describe the proposed approach which transfers knowledge between datasets in class IL without memory. 
We first propose a formalization of the problem and then introduce an adaptation of a prediction bias correction layer used in class IL with memory.
Finally, we introduce the knowledge transfer method which enables the use of the bias correction layer in class IL without memory. 

\subsection{Class-incremental learning formalization}
We adapt the class IL definition from~\cite{castro2018_e2eil, hou2019_lucir, rebuffi2017_icarl} to a setting without memory which includes a sequence of $S$ states. 
The first one is called initial state and the $S-1$ remaining states are incremental.
A set of $P_s$ new classes is learned in the $s^{th}$ state. 
IL states are disjoint and
\noindent $P_i \cap P_j = \varnothing\ \forall i,j \in \lb 1,S \rb, i \neq j$. 
A model $\mathcal{M}_1$ is initially trained on a dataset $\mathcal{D}_1 = \{ (X_1^j, Y_1^j) \colon j \in P_1 \}$, where $X_1^j$ and $Y_1^j$ are the sets of training images and their labels. 
We note $N_s$ the set of all classes seen until the $s^{th}$ state included. 
Thus, $N_1 = P_1$ initially, and $N_s = N_{s-1} \cup P_s = P_1  \cup  P_2  \cup ... \cup P_{s-1} \cup P_s$ for subsequent states. 
$\mathcal{M}_s$ is updated with an IL algorithm $\mathcal{A}$ using $\mathcal{D}_s = \{ (X_s^j, Y_s^j) \colon j \in P_s \}$. 
$\mathcal{D}_s$ includes only new classes samples, but $\mathcal{M}_s$ is evaluated on all classes seen so far ($j \in N_s$).
This makes the evaluation prone to catastrophic forgetting due to the lack of past exemplars~\cite{belouadah2021_study,masana2021_study}. 

\subsection{Adaptive bias correction layer}
\label{subsec:adbic}
\begin{figure}
    \centering

    \begin{subfigure}[b]{\columnwidth}
        \centering

        \hspace{-1.5em}
        \includegraphics[width=0.45\columnwidth]{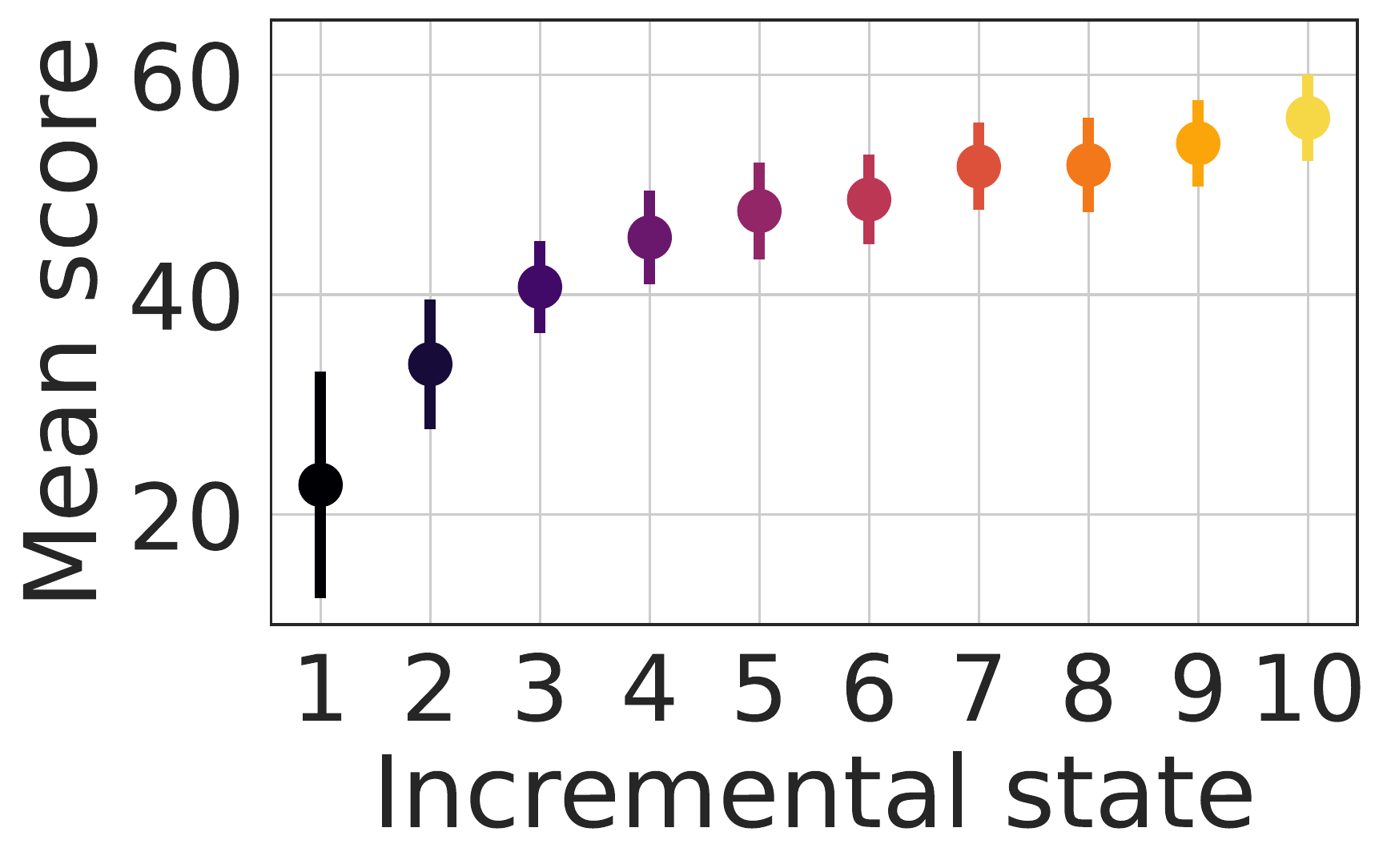}
        \hspace{0.25em}
        \includegraphics[width=0.45\columnwidth]{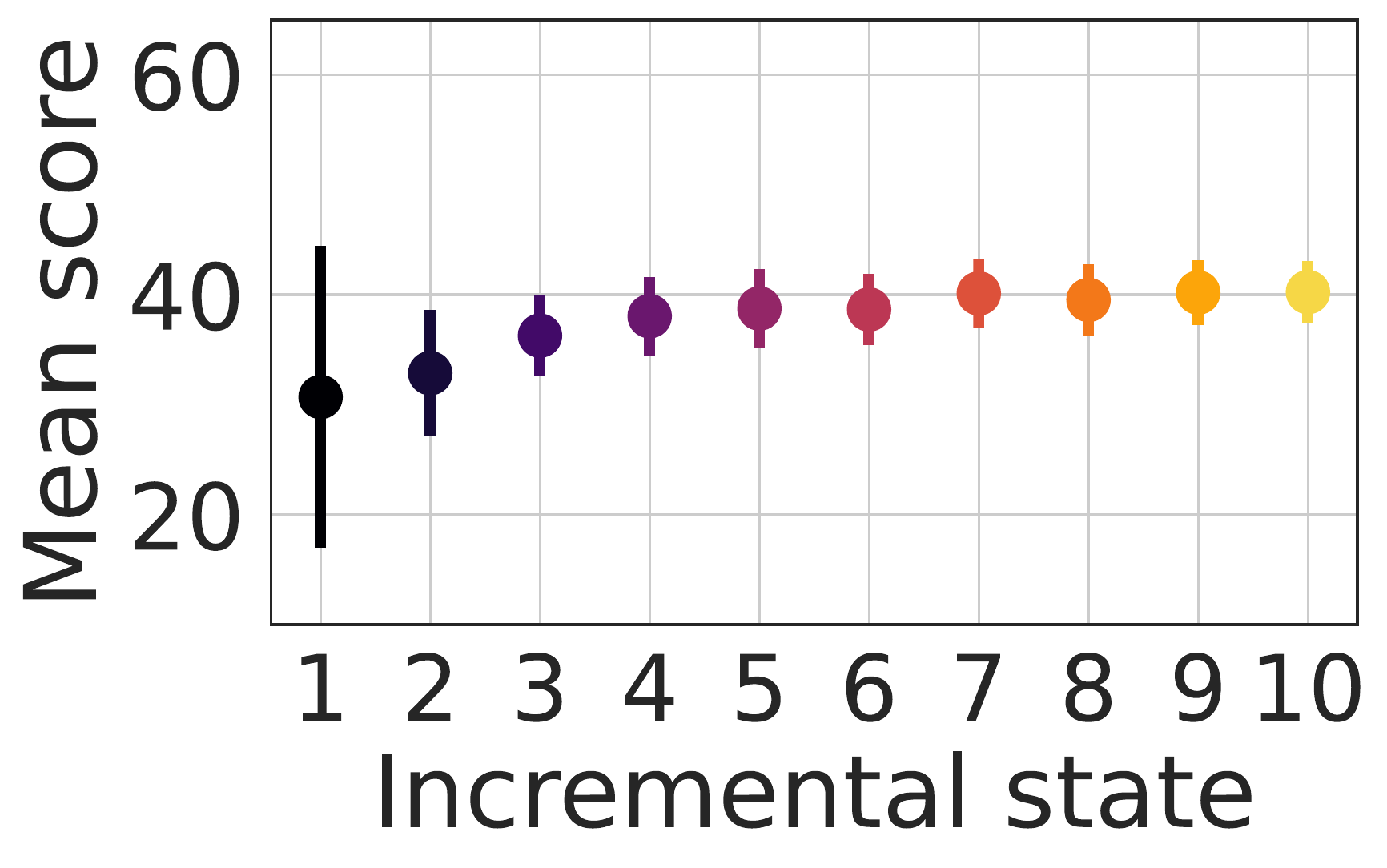}
        \caption{LUCIR \cite{hou2019_lucir}}
    \end{subfigure}
    
    \vspace{1em}
    
    \begin{subfigure}[b]{\columnwidth}
        \centering

        \hspace{-1.5em}
        \includegraphics[width=0.45\columnwidth]{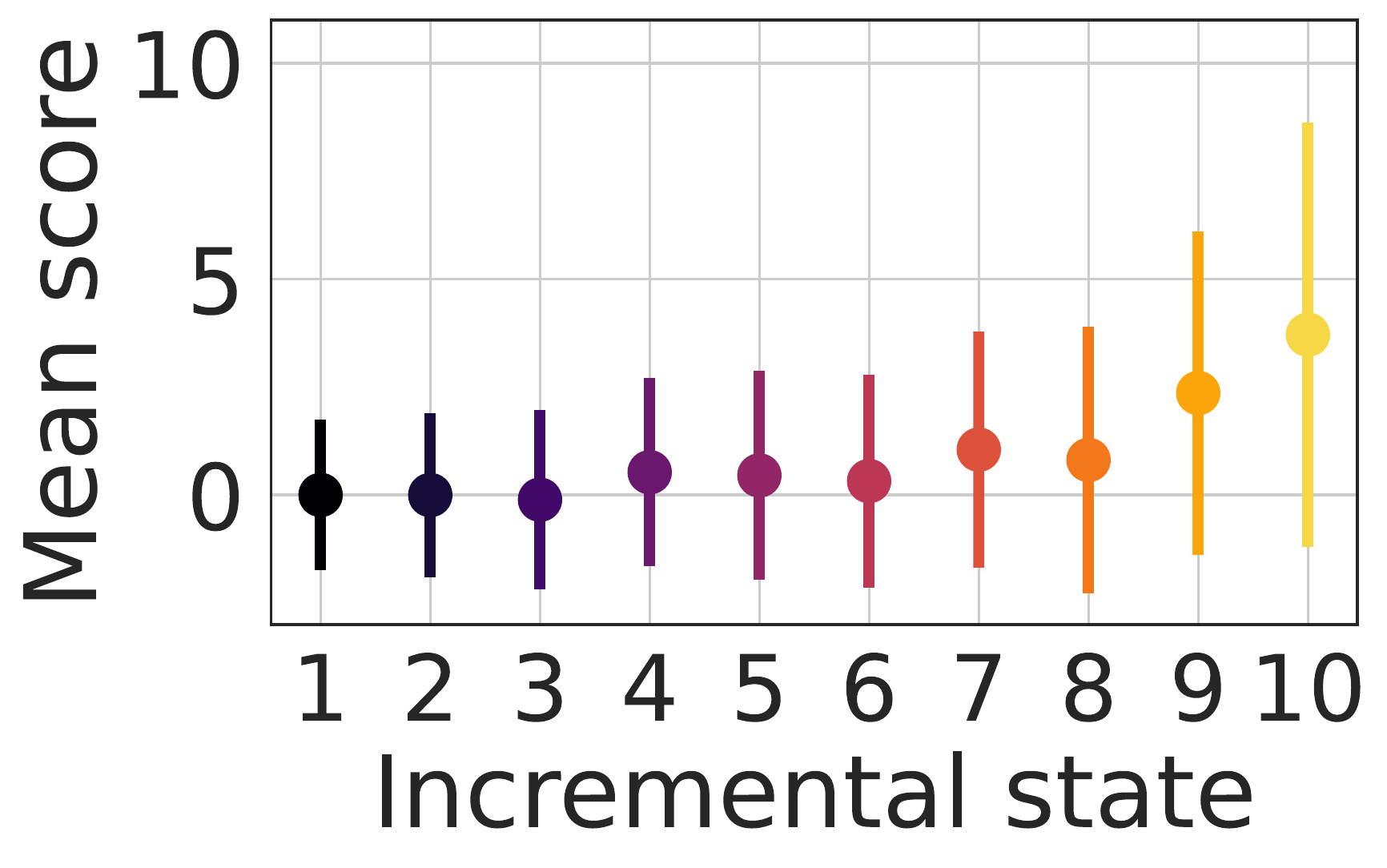}
        \hspace{0.25em}
        \includegraphics[width=0.45\columnwidth]{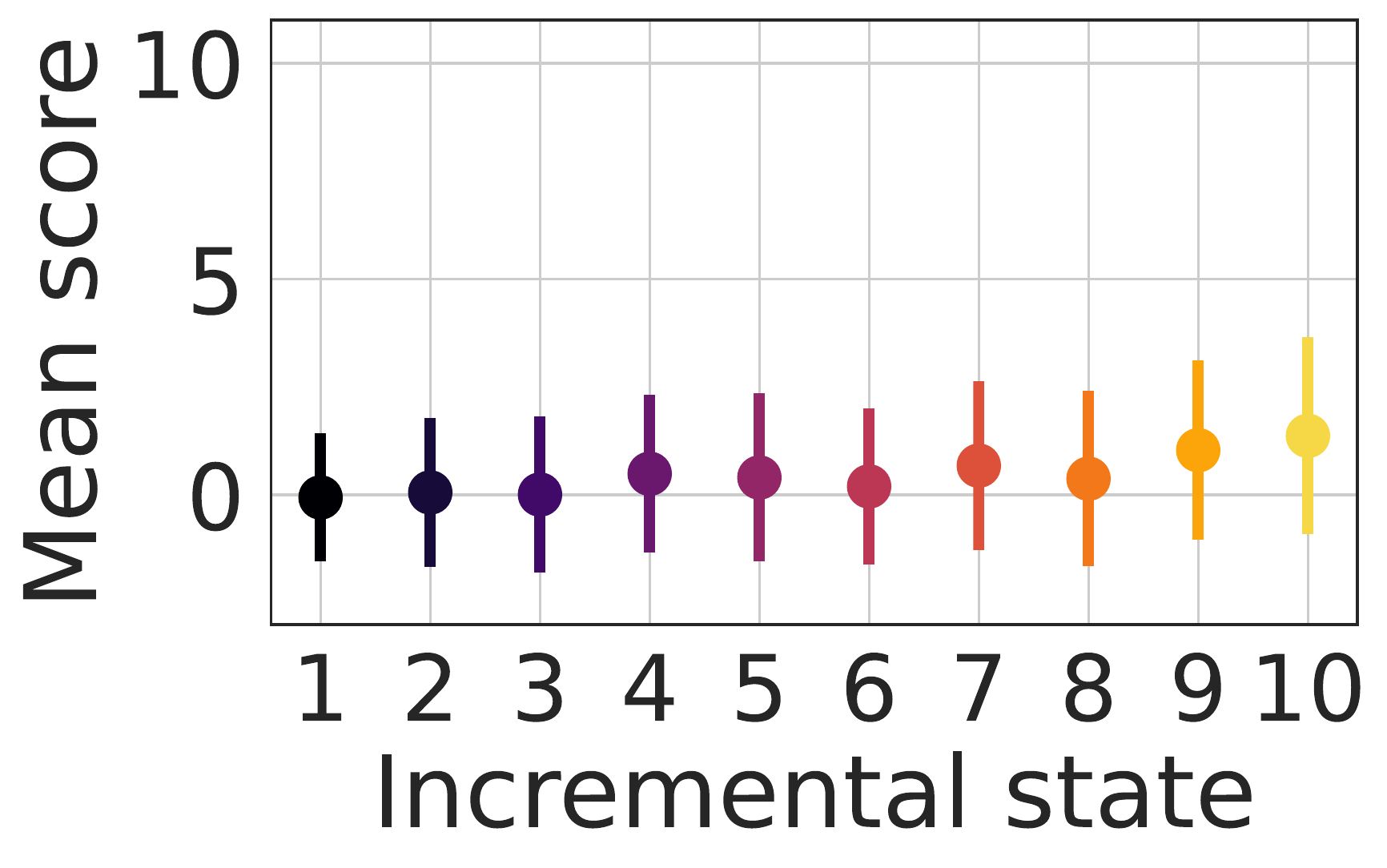}
        \caption{LwF \cite{li2016_lwf}}
    \end{subfigure}

    \caption{Mean prediction scores and standard deviations for \cifar classes grouped by state at the end of an IL process with $S=10$ states, for \lwf and \lucir, before (left) and after (right) calibration with $TransIL$.}
	\label{fig:mean_scores}
\end{figure}

The unavailability of past class exemplars when updating the incremental models leads to a classification bias toward new classes~\cite{wu2019_bic,zhao2020_maintaining}. 
We illustrate this in Figure~\ref{fig:mean_scores} (\textit{left}) by plotting mean prediction scores per state for the \cifar dataset with $S=10$ splits using \lucir and \lwf, the two distillation-based approaches tested here. 
Figure~\ref{fig:mean_scores} confirms that recently learned classes are favored, despite the use of knowledge distillation to counter the effects of catastrophic forgetting. 
New classes, learned in the last state, are particularly favored.
The predictions profiles for \lucir and \lwf are different.
\lucir mean predictions per state increase from earlier to latest states, while the tendency is less clear for \lwf.
\lwf predictions also have a stronger deviation in each state.
These observations make \lucir a better candidate for bias correction compared to \lwf.

Among the methods proposed to correct bias, the linear layer introduced in~\cite{wu2019_bic} is interesting for its simplicity and effectiveness~\cite{belouadah2021_study}.
This layer is defined in the $s^{th}$ state as:
\begin{align}\label{eq:bic}
    BiC(\bm{o_s^k}) =
\left\{
    \begin{array}{ll}
        \bm{o_s^k}
    & \mbox{if }~~ k \in \lb 1,\ s-1 \rb \\
        \alpha_s \bm{o_s^k} + \beta_s \cdot \mathbf{1}
    & \mbox{if }~~k=s
    \end{array}
\right.
\end{align}

where $\bm{o^k_s}$ are the raw scores of classes first seen in the $k^{th}$ state, obtained with $\mathcal{M}_s$; ($\alpha_s$, $\beta_s$) are the bias correction parameters in the $s^{th}$ state, and $\mathbf{1}$ is a vector of ones.

Equation~\ref{eq:bic} rectifies the raw predictions of new classes learned in the $s^{th}$ state to make them more comparable to those of past classes. 
The deep model is first updated using $\mathcal{D}_s$ containing new classes for this state.
The model is then frozen and calibration parameters ($\alpha_s$ and $\beta_s$) are optimized using a validation set made of samples of new and past classes. 
We remind that Equation~\ref{eq:bic} is not applicable in class IL without memory, the scenario explored here, because samples of past classes are not allowed. 
Figure~\ref{fig:mean_scores} (\textit{left}) shows that mean scores of classes learned in different incremental states are variable, which confirms that the amount of forgetting is uneven across past states. 
It is important to tune bias correction for classes which were learned in different IL states.
We thus define an adaptive version of \bic which rectifies predictions in the $s^{th}$ state with:

\begin{equation}
    adBiC(\bm{o_s^k}) = \alpha_s^k \bm{o_s^k} + \beta_s^k \cdot \mathbf{1} ~;\ \ k \in \lb 1, s \rb
\label{eq:adaptive_bic}
\end{equation}

where $\alpha_s^k$, $\beta_s^k$ are the parameters applied in the $s^{th}$ state to classes first learned in the $k^{th}$ state. 

Differently from Equation~\ref{eq:bic}, Equation~\ref{eq:adaptive_bic} adjusts prediction scores depending on the state in which classes were first encountered in the IL process. 
Note that each $\alpha_s^k$, $\beta_s^k$ pair is shared between all classes first learned in the same state.
These parameters are optimized on a validation set using the cross-entropy loss, defined for one data point $(\mathbf{x}, y)$ as:

\begin{equation}
    \mathcal{L} (\bm{q_s}, y)
    = - \sum_{k=1}^{s} \sum_{i=1}^{|P_k|}
        \delta_{y=\widehat{y}}
        \ \log\left(q_{s,i}^k\right)
\label{eq:adaptive_bic_optim}
\end{equation}

where $y$ is the ground-truth label, $\widehat{y}$ is the predicted label, $\delta$ is the Kronecker delta, and $\bm{q_s}$ is the softmax output for the sample corrected via Equation~\ref{eq:adaptive_bic}, defined as:
\begin{align}
    \bm{q_s}
        &= \upsigma
            \left(\left[
                \alpha_s^1 \bm{o_s^1} + \beta_s^1 \cdot \mathbf{1}
                \ \bm ;\ \cmdots\ \ \bm ;
                \alpha_s^s \bm{o_s^s} + \beta_s^s \cdot \mathbf{1}
            \right]\right)
\label{eq:corrected_output_def}
\end{align}

where $\upsigma$ is the softmax function. 

All $\alpha_s^k, \beta_s^k$ pairs are optimized using validation samples from classes in $N_s$.
We compare \adbic over \bic for our class IL setting in the evaluation section and show that the adaptation proposed here has a positive effect.

\subsection{Transferring knowledge between datasets}
The optimization of $\alpha$ and $\beta$ parameters is impossible in class IL without memory, since exemplars of past classes are unavailable. 
To circumvent this problem, we hypothesize that optimal values of these parameters can be transferred between reference and target datasets, noted $\mathcal{D}^r$ and $\mathcal{D}^t$ respectively. 
The intuition is that these values are sufficiently stable despite dataset content variability.
We create a set of reference datasets and perform a modified class IL training for them using the procedure described in Algorithm~\ref{algo:optimization}.
The modification consists in exploiting a validation set which includes exemplars of classes from all incremental states.
Validation set storage is necessary in order to optimize the parameters from Equation \ref{eq:adaptive_bic} and is possible since reference dataset training is done offline. 
Note that backbone incremental models for $\mathcal{D}^r$ are trained without memory in order to simulate the IL setting of target datasets $\mathcal{D}^t$.
We then store bias correction parameters optimized for reference datasets in order to perform transfer toward target datasets without using a memory.
For each incremental state, we compute the average of $\alpha$ and $\beta$ values over all reference datasets. The obtained averages are used for score rectification on target datasets.
This transfer uses the procedure described in Algorithm~\ref{algo:inference}.
The memory needed to store transferred parameters is negligible since we need $2 \times (2+3+...+S)=(S+2) \times (S-1)$ floats for each dataset and $S$ value. 
For $S=\{5,10,20\}$ states, we thus only store 28, 108 and 418 floating-point values respectively.

\begin{algorithm}
\label{algo:parameters}
\SetAlgoLined
\textbf{inputs :} $\mathcal{A}, \mathcal{D}_s^r$ for $s \in \lb 1, S \rb$ \Comment{\emph{reference dataset}} \\
randomly initialize $\mathcal{M}_1$ \;
$\mathcal{M}_1^* \leftarrow$ train($\mathcal{A}; \mathcal{M}_{1}, \mathcal{D}_1^r$) \;

\For{$s = 2 \cmdots S$}{
    $\mathcal{M}_s^* \leftarrow$ update($\mathcal{A}; \mathcal{M}^*_{s-1}, \mathcal{D}_s^r$) \;
    $\alpha_s^k \leftarrow 1,\ \beta_s^k \leftarrow 0\ $ for each $k \in \lb 1, s \rb$ \;
    
    \ForEach{$(\mathbf{x}, y) \in \mathcal{D}_s^r$ \Comment{validation set} \\}{ 
    $\mathbf{o_s} \leftarrow \mathcal{M}_s^*(\mathbf{x}) $ \; 
        \For{$k = 1 \cmdots s$}{
            $\mathbf{o_s^{k}}~\leftarrow~adBiC(\bm{o_s^k}) = \alpha_s^k \bm{o_s^k} + \beta_s^k \cdot \mathbf{1}$ \;
        }
        $\mathbf{q_s} \leftarrow \sigma(\mathbf{o_s})$ \;
        
        loss $\leftarrow \mathcal{L} (\bm{q_s}, y)$ \;
        $(\alpha_s^1, \beta_s^1, \cmdots, \alpha_s^s, \beta_s^s) \leftarrow $ optimize(loss) \;
    }
}
\caption{\small Optimization of calibration parameters}
\label{algo:optimization}
\end{algorithm}

\begin{algorithm}
    \label{algo:adbic}
    \SetAlgoLined
    \textbf{inputs :} $\mathcal{A}, (\alpha_s^k, \beta_s^k)$ averaged on reference datasets 
    \hspace*{3em} for each $s \in \lb 1, S \rb$, $k \in \lb 1, s \rb$\\
    \textbf{inputs :} $\mathcal{D}_s^t$ for $s \in \lb 1, S \rb$ \Comment{\emph{target dataset}} \\
    randomly initialize $\mathcal{M}_1$ \;
    $\mathcal{M}_1^* \leftarrow$ train($\mathcal{A}; \mathcal{M}_{1}, \mathcal{D}_1^t$)\; 
    
    \For{$s = 2 \cmdots S$}{
        $\mathcal{M}_s^* \leftarrow$ update($\mathcal{A}; \mathcal{M}^*_{s-1}, \mathcal{D}_s$) \; 
        
        \ForEach{$(\mathbf{x}, y) \in \mathcal{D}_s^t$ \Comment{test set} \\}{ 
        $\mathbf{o_s} \leftarrow \mathcal{M}_s^*(\mathbf{x}) $ \; 
        \For{$k = 1 \cmdots s$}{
            $\mathbf{o_s^{k}}~\leftarrow~adBiC(\bm{o_s^k}) = \alpha_s^k \bm{o_s^k} + \beta_s^k \cdot \mathbf{1}$ \;
            
        }
        $\mathbf{q_s} \leftarrow \sigma(\mathbf{o_s})$ \;
        $\hat{y} \leftarrow \underset{y \in \lb 1, N_s \rb}{argmax}(\mathbf{q_s})$ ; \Comment{\emph{inference}}  \\
        
        }
    }
    \caption{\adbic inference}
    \label{algo:inference}
\end{algorithm}

\begin{figure}
    \centering
    \begin{subfigure}{\columnwidth}
        \centering
        \includegraphics[width=\linewidth]{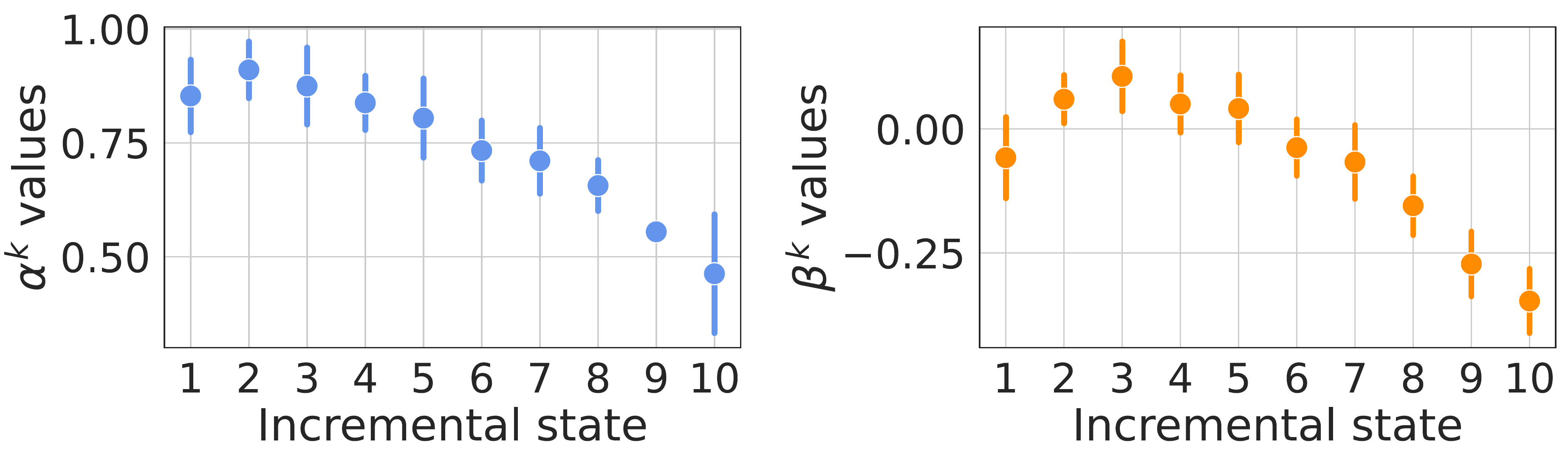}
        \vspace{-15pt}
        \caption{LwF \cite{li2016_lwf}}
    \end{subfigure}

    \vspace{0.5em}
    \begin{subfigure}{\columnwidth}
        \centering
        \includegraphics[width=\linewidth]{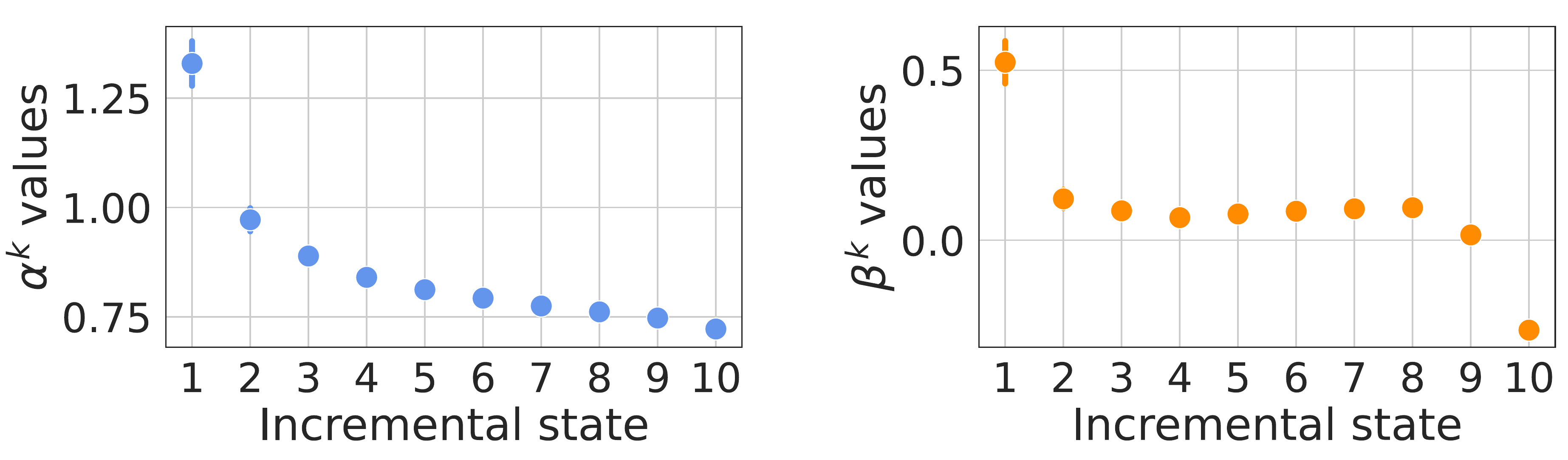}
        \vspace{-15pt}
        \caption{LUCIR \cite{hou2019_lucir}}
    \end{subfigure}
    \vspace{-1mm}
    \caption{Averaged $\alpha^k$ (left) and $\beta^k$ (right) values computed for $R=10$ reference datasets using \lwf and \lucir, at the end of an incremental process with $S=10$ states.}
    \vspace{-4mm}
	\label{fig:range_analysis}
\end{figure}

In Figure~\ref{fig:range_analysis}, we illustrate optimal parameters obtained across $R=10$ reference datasets which are further described in Section~\ref{sec:eval}.
We plot $\alpha^k$ and $\beta^k$ values learned after $S=10$ IL states, using \lwf~\cite{li2016_lwf} and \lucir~\cite{hou2019_lucir} methods.
Mean and standard deviations are presented for past and current incremental states in the final state of the IL process.
The parameter ranges from Figure~\ref{fig:range_analysis} confirm that, while optimal values do vary across datasets, this variation is rather low and calibration profiles remain similar.
Consequently, parameters are transferable.
When $R > 1$, a transfer function is needed to apply the parameters learned on reference datasets to a target dataset. 
We transfer parameters using the averaged $\alpha_s^k$ and $\beta_s^k$ values, obtained for the set of $\mathcal{D}^r$.
In Section~\ref{sec:eval}, we evaluate this transfer against an upper-bound oracle which selects the best $\mathcal{D}^r$ in each state.

The proposed approach adds a simple but effective linear layer to calibrate the predictions of backbone class IL methods.
Consequently, it is applicable to any IL method which works without memory. 
We test the genericity of the approach by applying it on top of four existing methods. 

\section{Evaluation}
\label{sec:eval}
In this section, we discuss: 
(1) the reference and target datasets, (2) the backbone methods to which bias correction is applied and (3) the analysis of the obtained results.
The evaluation metric is the average top-1 accuracy of the IL process introduced in~\cite{rebuffi2017_icarl}, which combines the accuracy obtained for individual incremental states.
Following~\cite{castro2018_e2eil}, we discard the accuracy of the first state since it is not incremental.
We use a ResNet-18 backbone whose implementation details are provided in the supp. material.

\subsection{Datasets}
\textbf{Reference datasets.}
The preliminary analysis from Figure~\ref{fig:range_analysis} indicates that bias correction parameters are rather stable for different reference datasets.
It is interesting to use several such datasets in order to stabilize averaged bias correction parameters.
In our experiments, we use 10 reference datasets, each including 100 randomly chosen leaf classes from ImageNet~\cite{deng2009_imagenet}, with a 500/200 train/val split per class.
There is no intersection between these datasets, as each class appears only in one of them.

\textbf{Target datasets.}
We test \textit{TransIL}  with four target datasets.
They were selected to include different types of visual content and thus test the robustness of the parameter transfer.
The class samples from the target datasets are split into 500/100 train/test subsets respectively. 
There is no intersection between the classes from the reference datasets and the two target datasets which are sampled from ImageNet. 
We describe target datasets briefly hereafter and provide details in the supplementary material: 
\noindent\ \ $\sbullet[.75] ~$ \cifar \cite{krizhevsky2009_cifar100} - object recognition dataset. It focuses on commonsense classes and is relevant for basic level classification in the sense of ~\cite{rosch1999_principles}.

\noindent\ \ $\sbullet[.75] ~$ \imnet\ - subset of ImageNet~\cite{deng2009_imagenet} which includes 100 randomly selected leaf classes. It is built with the same procedure used for reference datasets and is thus most similar to them. \imnet is relevant for fine-grained classification with a diversity of classes.

\noindent\ \ $\sbullet[.75] ~$ \birds\ - uses 100 bird classes from ImageNet~\cite{deng2009_imagenet}. It is built for domain fine-grained classification.

\noindent\ \ $\sbullet[.75] ~$ \food\ - uses 100 food classes from Food-101~\cite{bossard2014_food101}. It is a fine-grained and domain-specific dataset and is interesting because it is independent from ImageNet.

\subsection{Backbone incremental learning methods}
We apply \adbic on top of four backbone methods which are usable for class IL without memory:

\noindent\ \ $\sbullet[.75] ~$ \lwf~\cite{rebuffi2017_icarl} - version of the original method from~\cite{li2016_lwf} which exploits distillation to reduce catastrophic forgetting for past classes.

\noindent\ \ $\sbullet[.75] ~$ \lucir~\cite{hou2019_lucir} - distillation-based approach which uses a more elaborate way of ensuring a good balance between model stability and plasticity. We use the CNN version because it is adaptable to our setting.

\noindent\ \ \ $\sbullet[.75] ~$\ \ftplus~\cite{masana2021_study} - fine-tuning in which past classes weights are not updated to reduce catastrophic forgetting.

\noindent\ \ $\sbullet[.75] ~$ \siw~\cite{belouadah2020_siw} - similar to \ftplus, but with class weights standardization added to improve the comparability of prediction between past and new classes.

We compare \adbic to \bic, the original linear layer from~\cite{wu2019_bic}. 
We also provide results with an optimal version of \adbic, which is obtained via an oracle-based selection of the best performing reference dataset for each IL state.
This oracle is important as it indicates the potential supplementary gain obtainable with a parameter selection method more refined than the proposed one.
Finally, we provide results with \textit{Joint}, a training from scratch with all data available at all times.
This is an upper bound for all IL methods.

\begin{table*}[t]
    \begin{center}
    \resizebox{\linewidth}{!}{
    \begin{tabular}{@{\kern0.5em}lccccccccccccccc@{\kern0.5em}}
        \toprule
        \multirow{2}{*}{\textbf{Method}}
            & \multicolumn{3}{c}{\cifar}
            && \multicolumn{3}{c}{\imnet}
            && \multicolumn{3}{c}{\birds}
            && \multicolumn{3}{c}{\food}
        \\ \cmidrule(lr){2-4} \cmidrule(lr){6-8} \cmidrule(l){10-12}  \cmidrule(l){14-16}
        
            & \multicolumn{1}{c}{\textit{S}\ =\ 5}
            & \multicolumn{1}{c}{\textit{S}\ =\ 10}
            & \multicolumn{1}{c}{\textit{S}\ =\ 20}
            &
            & \multicolumn{1}{c}{\textit{S}\ =\ 5}
            & \multicolumn{1}{c}{\textit{S}\ =\ 10}
            & \multicolumn{1}{c}{\textit{S}\ =\ 20}
            &
            & \multicolumn{1}{c}{\textit{S}\ =\ 5}
            & \multicolumn{1}{c}{\textit{S}\ =\ 10}
            & \multicolumn{1}{c}{\textit{S}\ =\ 20}
            &
            & \multicolumn{1}{c}{\textit{S}\ =\ 5}
            & \multicolumn{1}{c}{\textit{S}\ =\ 10}
            & \multicolumn{1}{c}{\textit{S}\ =\ 20}
        \\ \midrule \midrule

        \addlinespace[0.25em]
        \textbf{LwF} \cite{li2016_lwf}                           
        & 53.0 & 44.0 & 29.1 && 53.8 & 41.1 & 29.2 && 53.7 & 41.8 & 30.1  && 42.9 & 31.8 & 22.2 \\
        \textit{w}/ \bic    
        & 54.0 \tabnoteg{+ 1.0} & 45.5 \tabnoteg{+ 1.5} & 30.8 \tabnoteg{+ 1.7} &&
          54.7 \tabnoteg{+ 0.9} & 42.5 \tabnoteg{+ 1.4} & 31.1 \tabnoteg{+ 1.9} &&
          54.6 \tabnoteg{+ 0.9} & 43.1 \tabnoteg{+ 1.3} & 31.8 \tabnoteg{+ 1.7} &&
          43.4 \tabnoteg{+ 0.5} & 32.6 \tabnoteg{+ 0.8} & 23.8 \tabnoteg{+ 1.6} \\

        \rowcolor[RGB]{240,240,240}
        \textit{w}/ \adbic    
        & 54.3 \tabnoteg{+ 1.3} & \textbf{46.4} \tabnoteg{+ 2.4} & \textbf{32.3} \tabnoteg{+ 3.2} &&
          55.1 \tabnoteg{+ 1.3} & 43.4 \tabnoteg{+ 2.3} & \textbf{32.3} \tabnoteg{+ 3.1} &&
          55.0 \tabnoteg{+ 1.3} & 44.0 \tabnoteg{+ 2.2} & \textbf{32.8} \tabnoteg{+ 2.7} &&
          43.5 \tabnoteg{+ 0.6} & 33.3 \tabnoteg{+ 1.5} & \textbf{24.7} \tabnoteg{+ 2.5} \\

        \textit{w}/ \adbic + $\mathbb{O}$
        & 54.9 \tabnoteg{+ 1.9} & 47.3 \tabnoteg{+ 3.3} & 32.6 \tabnoteg{+ 3.5} &&
          55.9 \tabnoteg{+ 2.1} & 44.2 \tabnoteg{+ 3.1} & 33.1 \tabnoteg{+ 3.9} &&
          55.8 \tabnoteg{+ 2.1} & 44.8 \tabnoteg{+ 3.0} & 33.3 \tabnoteg{+ 3.2} &&
          44.0 \tabnoteg{+ 1.1} & 34.2 \tabnoteg{+ 2.4} & 25.3 \tabnoteg{+ 3.1} \\
      \midrule
        \addlinespace[0.25em]

        \textbf{LUCIR} \cite{hou2019_lucir}
        & 50.1 & 33.7 & 19.5 && 48.3 & 30.1 & 17.7 && 50.8 & 31.4 & 17.9  && 44.2 & 26.4 & 15.5 \\
        \textit{w}/ \bic    
        & 52.5 \tabnoteg{+ 2.4} & 37.1 \tabnoteg{+ 3.4} & 22.4 \tabnoteg{+ 2.9} &&
          54.9 \tabnoteg{+ 6.6} & 36.8 \tabnoteg{+ 6.7} & 21.8 \tabnoteg{+ 4.1} &&
          56.0 \tabnoteg{+ 5.2} & 37.7 \tabnoteg{+ 6.3} & 20.6 \tabnoteg{+ 2.7} &&
          49.9 \tabnoteg{+ 5.7} & 31.5 \tabnoteg{+ 5.1} & 17.2 \tabnoteg{+ 1.7} \\

        \rowcolor[RGB]{240,240,240}
        \textit{w}/ \adbic    
        & \textbf{54.8} \tabnoteg{+ 4.7}  & 42.2 \tabnoteg{+ 8.5}  & 28.4 \tabnoteg{+ 8.9} &&
          \textbf{59.0} \tabnoteg{+ 10.7} & \textbf{46.1} \tabnoteg{+ 16.0} & 27.3 \tabnoteg{+ 9.6} &&
          \textbf{58.5} \tabnoteg{+ 7.7}  & \textbf{45.4} \tabnoteg{+ 14.0} & 27.3 \tabnoteg{+ 9.4} &&
          \textbf{52.0} \tabnoteg{+ 7.8}  & \textbf{37.1} \tabnoteg{+ 10.7} & 17.7 \tabnoteg{+ 2.2} \\

        \textit{w}/ \adbic + $\mathbb{O}$
        & 55.5 \tabnoteg{+ 5.4}  & 43.6 \tabnoteg{+ 9.9}  & 31.2 \tabnoteg{+ 11.7} &&
          59.4 \tabnoteg{+ 11.1} & 46.6 \tabnoteg{+ 16.5} & 29.7 \tabnoteg{+ 12.0} &&
          59.0 \tabnoteg{+ 8.2}  & 46.0 \tabnoteg{+ 14.6} & 28.8 \tabnoteg{+ 10.9} &&
          52.6 \tabnoteg{+ 8.4}  & 38.2 \tabnoteg{+ 11.8} & 21.0 \tabnoteg{+ 5.5} \\
        \midrule
        \addlinespace[0.25em]

        \textbf{SIW} \cite{belouadah2020_siw}                          
        & 29.9 & 22.7 & 14.8 && 32.6 & 23.3 & 15.1 && 30.6 & 23.2 & 14.9 && 29.4 & 21.6 & 14.1 \\
        \textit{w}/ \bic    
        & 31.4 \tabnoteg{+ 1.5}  & 22.8 \tabnoteg{+ 0.1} & 14.7 \tabnoter{- 0.1} &&
          33.9 \tabnoteg{+ 1.3}  & 22.6 \tabnoter{- 0.7} & 13.9 \tabnoter{- 1.2} &&
          32.8 \tabnoteg{+ 2.2}  & 22.7 \tabnoter{- 0.5} & 12.8 \tabnoter{- 2.1} &&
          29.1 \tabnoter{- 0.3}  & 20.3 \tabnoter{- 1.3} & 12.1 \tabnoter{- 2.0} \\

        \rowcolor[RGB]{240,240,240}
        \textit{w}/ \adbic
        & 31.7 \tabnoteg{+ 1.8} & 24.1 \tabnoteg{+ 1.4} & 15.8 \tabnoteg{+ 1.0} &&
          35.1 \tabnoteg{+ 2.5} & 24.5 \tabnoteg{+ 1.2} & 15.0 \tabnoter{- 0.1} &&
          33.0 \tabnoteg{+ 2.4} & 25.2 \tabnoteg{+ 2.0} & 15.3 \tabnoteg{+ 0.4} &&
          30.9 \tabnoteg{+ 1.5} & 21.3 \tabnoter{- 0.3} & 14.5 \tabnoteg{+ 0.4} \\

        \textit{w}/ \adbic + $\mathbb{O}$
        & 32.8 \tabnoteg{+ 2.9} & 25.0 \tabnoteg{+ 2.3} & 16.5 \tabnoteg{+ 1.7} &&
          36.4 \tabnoteg{+ 3.8} & 25.7 \tabnoteg{+ 2.4} & 16.1 \tabnoteg{+ 1.0} &&
          34.4 \tabnoteg{+ 3.8} & 26.2 \tabnoteg{+ 3.0} & 16.3 \tabnoteg{+ 1.4} &&
          31.5 \tabnoteg{+ 2.1} & 22.6 \tabnoteg{+ 1.0} & 15.1 \tabnoteg{+ 1.0} \\
        \midrule
        \addlinespace[0.25em]

        \textbf{FT+}
        & 28.9 & 22.6 & 14.5 && 31.7 & 23.2 & 14.6 && 29.7 & 23.3 & 13.5  && 28.7 & 21.1 & 13.3 \\
        \textit{w}/ \bic    
        & 30.7 \tabnoteg{+ 1.8} & 22.5 \tabnoter{- 0.1} & 14.8 \tabnoteg{+ 0.3} &&
          33.0 \tabnoteg{+ 1.3} & 21.9 \tabnoter{- 1.3} & 13.8 \tabnoter{- 0.8} &&
          32.3 \tabnoteg{+ 2.6} & 22.5 \tabnoter{- 0.8} & 12.4 \tabnoter{- 1.1} &&
          28.6 \tabnoter{- 0.1} & 20.6 \tabnoter{- 0.5} & 11.8 \tabnoter{- 1.5} \\

        \rowcolor[RGB]{240,240,240}
        \textit{w}/ \adbic    
        & 31.9 \tabnoteg{+ 3.0} & 23.6 \tabnoteg{+ 1.0} & 15.0 \tabnoteg{+ 0.5} &&
          34.9 \tabnoteg{+ 3.2} & 23.7 \tabnoteg{+ 0.5} & 15.7 \tabnoteg{+ 1.1} &&
          34.0 \tabnoteg{+ 4.3} & 25.0 \tabnoteg{+ 1.7} & 14.2 \tabnoteg{+ 0.7} &&
          30.8 \tabnoteg{+ 2.1} & 22.2 \tabnoteg{+ 1.1} & 14.2 \tabnoteg{+ 0.9} \\

        \textit{w}/ \adbic + $\mathbb{O}$
        & 32.5 \tabnoteg{+ 3.6} & 24.6 \tabnoteg{+ 2.0} & 15.9 \tabnoteg{+ 1.4} &&
          35.7 \tabnoteg{+ 4.0} & 24.9 \tabnoteg{+ 1.7} & 16.2 \tabnoteg{+ 1.6} &&
          34.5 \tabnoteg{+ 4.8} & 25.7 \tabnoteg{+ 2.4} & 15.4 \tabnoteg{+ 1.9} &&
          31.3 \tabnoteg{+ 2.6} & 22.7 \tabnoteg{+ 1.6} & 14.5 \tabnoteg{+ 1.2} \\
        \hline \hline
        \addlinespace[0.25em]

        \textit{Joint}
                & \multicolumn{3}{c}{72.7}
                && \multicolumn{3}{c}{75.5}
                && \multicolumn{3}{c}{80.9} 
                && \multicolumn{3}{c}{71.03} \\

        \bottomrule

        \end{tabular}
    }
    \end{center}
    \vspace{-4mm}
    \caption{Average top-1 incremental accuracy using $S=\{5,10,20\}$ states. Results are presented for each method without parameter transfer and with \bic and \adbic transfer. The gain (green) and loss (red) in accuracy obtained with parameter transfer are provided for each configuration. \textit{Joint} is an upper bound obtained using a standard training with all data available. $\mathbb{O}$ denotes a choice of the reference dataset by oracle, in which the best reference dataset for each state is selected for transfer. Best results for each setting (excluding the oracle) are in bold. A graphical view of this table is in the supplementary material.}
    \vspace{-1mm}
    \label{tab:global}
\end{table*}

\subsection{Overall results}

Results from Figure \ref{fig:mean_scores} (\textit{right}) indicate that the degree of forgetting depends on the initial state in which classes were first learned. Applying calibration parameters learned on reference datasets clearly reduces the imbalance of mean prediction scores and the bias toward recent classes.

Results from Table~\ref{tab:global} show that our method improves the performance of baseline methods for all but two of the configurations evaluated.
The best overall performance before bias correction is obtained with \lwf.
This result confirms the conclusions of~\cite{belouadah2020_siw,masana2021_study} regarding the strong performance of \lwf in class IL without memory for medium-scale datasets. 
With \adbic, \lucir performs generally better than \lwf for $S=5$ and $S=10$, while \lwf remains stronger with $S=20$ states.
Results are particularly interesting for \lucir, a method for which \adbic brings consistent gains (up to 16 accuracy points) in most configurations. 
Table~\ref{tab:global} shows that \adbic also improves the results of \lwf in all configurations, albeit to a lesser extent compared to \lucir.
Interestingly, improvements for \lwf are larger for $S=20$ states. 
This is the most challenging configuration since the model is more prone to forgetting.
\ftplus~\cite{masana2021_study} and \siw~\cite{belouadah2020_siw} remove the distillation component for the class IL training process and exploit the weights of past classes learned in their initial state.
\adbic improves results for these two methods in all but one configuration.
However, their global performance is significantly lower than that of \lwf and \lucir, the two methods which make use of distillation. 
This result confirms the finding from~\cite{belouadah2020_siw} regarding the usefulness of the distillation term exploited by \lwf and \lucir to stabilize IL training for medium scale datasets. 

Results from Table~\ref{tab:global} highlight the effectiveness of \adbic compared to \bic.
\adbic has better accuracy in all tested configurations, with the most important gain over \bic obtained for \lucir. 
It is also worth noting that \adbic improves results for \siw and \ftplus in most configurations, while the corresponding results of \bic are mixed. 
The  comparison of \adbic and \bic validates our hypothesis that a finer-grained modeling of forgetting for past states is better compared to a uniform processing of them.
It would be interesting to test the usefulness of \adbic in the class IL with memory setting originally tested in~\cite{wu2019_bic}. 

We also compare \adbic, which uses averaged $\alpha$ and $\beta$ parameters, with an oracle selection of parameters (+ $\mathbb{O}$).
The performance of \adbic is close to this upper bound for all tested methods.
This indicates that averaging parameters is an effective way to aggregate parameters learned from reference datasets. 
However, it would be interesting to investigate more refined ways to transfer parameters from reference to target datasets to further improve performance. 

The comparison of target datasets shows that the gain brought by \adbic is the largest for \imnet, followed by \birds, \cifar and \food. 
This is intuitive as \imnet has the closest distribution to that of reference datasets. 
\birds is extracted from ImageNet and, while topically different from reference datasets, was created using similar guidelines. 
The consistent improvements obtained with \cifar and \food, two datasets independent from ImageNet, shows that the proposed transfer method is robust to data distribution changes.
The performance gaps between IL results and $Joint$ are still wide, particularly for larger values of $S$.
This indicates that class IL without memory remains an open challenge.

Except for \lwf, \adbic gains are larger for $S=\{5,10\}$ compared to $S=20$.
This result is consistent with past findings reported for bias correction methods~\cite{masana2021_study,wu2019_bic}.
It is mainly explained by the fact that the size of validation sets needed to optimize \adbic parameters is smaller and thus less representative for larger values of $S$.
A larger number of states leads to a higher degree of forgetting.
This makes the IL training process more challenging and also has a negative effect on the usefulness of the bias correction layer.

Figure~\ref{fig:mean_scores} provides a qualitative view of the effect of \adbic for \lwf and \lucir which complements numerical results from Table~\ref{tab:global}.
The correction is effective since the predictions associated to IL states are more balanced (right), compared to the raw predictions (left). 
The effect of calibration is particularly interesting for \lucir, where mean prediction scores are balanced for states 3 to 10. 
We note that bias correction should ideally provide fully balanced mean prediction scores to give equal chances to classes learned in different states.
Some variation subsists and is notably due to variable forgetting for past states and to the variable difficulty of learning different visual classes.

\begin{table*}[h]
    \begin{center}
    \resizebox{0.99\linewidth}{!}{
    \begin{tabular}{@{\kern0.5em}lccccccccccccccc@{\kern0.5em}}
        \toprule
        \multirow{2}{*}{\textbf{Method}}
            & \multicolumn{3}{c}{\cifar}
            && \multicolumn{3}{c}{\imnet}
            && \multicolumn{3}{c}{\birds}
            && \multicolumn{3}{c}{\food}
        \\ \cmidrule(lr){2-4} \cmidrule(lr){6-8} \cmidrule(l){10-12} \cmidrule(l){14-16}
        
            & \multicolumn{1}{c}{\textit{S}\ =\ 5}
            & \multicolumn{1}{c}{\textit{S}\ =\ 10}
            & \multicolumn{1}{c}{\textit{S}\ =\ 20}
            &
            & \multicolumn{1}{c}{\textit{S}\ =\ 5}
            & \multicolumn{1}{c}{\textit{S}\ =\ 10}
            & \multicolumn{1}{c}{\textit{S}\ =\ 20}
            &
            & \multicolumn{1}{c}{\textit{S}\ =\ 5}
            & \multicolumn{1}{c}{\textit{S}\ =\ 10}
            & \multicolumn{1}{c}{\textit{S}\ =\ 20}
            &
            & \multicolumn{1}{c}{\textit{S}\ =\ 5}
            & \multicolumn{1}{c}{\textit{S}\ =\ 10}
            & \multicolumn{1}{c}{\textit{S}\ =\ 20}
        \\ \midrule \midrule

        \addlinespace[0.25em]
        \textbf{LwF} \cite{li2016_lwf}
        & 41.3 & 33.3 & 23.3 && 45.6 & 33.5 & 23.8 && 44.6 & 34.0 & 23.2  && 29.5 & 23.3 & 17.3 \\
        \rowcolor[RGB]{240,240,240}
        \textit{w}/ \adbic    
        & 42.1 \tabnoteg{+ 0.8} & 34.8 \tabnoteg{+ 1.5} & 25.0 \tabnoteg{+ 1.7} &&
          46.7 \tabnoteg{+ 1.1} & 35.3 \tabnoteg{+ 1.8} & 25.6 \tabnoteg{+ 1.8} &&
          45.5 \tabnoteg{+ 0.9} & 35.4 \tabnoteg{+ 1.4} & 25.2 \tabnoteg{+ 2.0} &&
          29.9 \tabnoteg{+ 0.4} & 24.3 \tabnoteg{+ 1.0} & 18.7 \tabnoteg{+ 1.4} \\
        \midrule

        \addlinespace[0.25em]
        \textbf{LUCIR} \cite{hou2019_lucir}
        & 43.5 & 27.8 & 16.6 && 42.9 & 27.6 & 17.0 && 45.2 & 27.8 & 16.0  && 37.9 & 22.7 & 13.9 \\
        \rowcolor[RGB]{240,240,240}
        \textit{w}/ \adbic
        & 48.3 \tabnoteg{+ 4.8} & 38.5 \tabnoteg{+ 10.7} & 25.3 \tabnoteg{+ 8.7} &&
          54.1 \tabnoteg{+ 11.2} & 42.4 \tabnoteg{+ 14.8} & 23.2 \tabnoteg{+ 6.2} &&
          52.8 \tabnoteg{+ 7.6} & 40.9 \tabnoteg{+ 13.1} & 25.6 \tabnoteg{+ 9.6} &&
            45.7 \tabnoteg{+ 7.8} & 32.6 \tabnoteg{+ 9.9} & 19.8 \tabnoteg{+ 5.9} \\
        \midrule

        \addlinespace[0.25em]
        \textbf{SIW} \cite{belouadah2020_siw}                          
        & 31.7 & 21.6 & 13.7 && 32.1 & 22.7 & 14.4 && 29.7 & 22.8 & 14.1  && 28.4 & 18.7 & 13.5 \\
        \rowcolor[RGB]{240,240,240}
        \textit{w}/ \adbic
        & 33.7 \tabnoteg{+ 2.0} & 22.5 \tabnoteg{+ 0.9} & 14.0 \tabnoteg{+ 0.3} &&
          35.0 \tabnoteg{+ 2.9} & 22.6 \tabnoter{- 0.1} & 12.2 \tabnoter{- 2.2} &&
          32.1 \tabnoteg{+ 2.4} & 23.7 \tabnoteg{+ 0.9} & 13.5 \tabnoter{- 0.6} &&
          29.9 \tabnoteg{+ 1.5} & 16.9 \tabnoter{- 1.8} & 13.3 \tabnoter{- 0.2} \\
        \midrule

        \addlinespace[0.25em]
        \textbf{FT+}
        & 30.4 & 21.5 & 12.9 && 31.2 & 22.2 & 12.0 && 29.2 & 22.8 & 12.2  && 27.4 & 18.2 & 11.6 \\
        \rowcolor[RGB]{240,240,240}
        \textit{w}/ \adbic    
        & 32.0 \tabnoteg{+ 1.6} & 21.4 \tabnoter{- 0.1} & 13.4 \tabnoteg{+ 0.5} &&
          34.8 \tabnoteg{+ 3.6} & 21.2 \tabnoter{- 1.0} & 13.7 \tabnoteg{+ 1.7} &&
          31.9 \tabnoteg{+ 2.7} & 23.0 \tabnoteg{+ 0.2} & 13.6 \tabnoteg{+ 1.4} &&
          28.8 \tabnoteg{+ 1.4} & 16.2 \tabnoter{- 2.0} & 12.2 \tabnoteg{+ 0.6} \\
        \bottomrule

        \end{tabular}
    }
    \end{center}
    \vspace{-5mm}
    \caption{Average top-1 IL accuracy with 50\% of training images for target datasets. Gains are in green, losses are in red. 
    }
    \vspace{-1mm}
    \label{tab:ablation_train}
\end{table*}

\begin{table*}[t]
    \begin{center}
    \resizebox{0.99\linewidth}{!}{
    \begin{tabular}{@{\kern0.5em}cccccccccccc@{\kern0.5em}}
\toprule

\multirow{2}{*}{\textbf{\textit{S}\ =\ 5}}  & Raw & \textit{R}\ =\ 1 & \textit{R}\ =\ 2 & \textit{R}\ =\ 3 & \textit{R}\ =\ 4 & \textit{R}\ =\ 5 & \textit{R}\ =\ 6 & \textit{R}\ =\ 7 & \textit{R}\ =\ 8 & \textit{R}\ =\ 9 & \textit{R}\ =\ 10 \\ \cmidrule(l){2-12} &
        
44.19 & 51.9 \tabnoteb{$\pm$ 0.4} & 52.0 \tabnoteb{$\pm$ 0.2} & 52.1 \tabnoteb{$\pm$ 0.2} & 52.0 \tabnoteb{$\pm$ 0.1} & 52.1 \tabnoteb{$\pm$ 0.1} & 52.0 \tabnoteb{$\pm$ 0.1} & 52.0 \tabnoteb{$\pm$ 0.1} & 52.0 \tabnoteb{$\pm$ 0.1} & 52.0 \tabnoteb{$\pm$ 0.1} &  
52.0
\\ \midrule \midrule

\multirow{2}{*}{\textbf{\textit{S}\ =\ 10}} & Raw & \textit{R}\ =\ 1 & \textit{R}\ =\ 2 & \textit{R}\ =\ 3 & \textit{R}\ =\ 4 & \textit{R}\ =\ 5 & \textit{R}\ =\ 6 & \textit{R}\ =\ 7 & \textit{R}\ =\ 8 & \textit{R}\ =\ 9 & \textit{R}\ =\ 10 \\ \cmidrule(l){2-12} &

26.44 & 36.7 \tabnoteb{$\pm$ 0.7} & 36.9 \tabnoteb{$\pm$ 0.4} & 37.2 \tabnoteb{$\pm$ 0.4} & 37.2 \tabnoteb{$\pm$ 0.3} & 37.1 \tabnoteb{$\pm$ 0.2} & 37.0 \tabnoteb{$\pm$ 0.2} & 37.0 \tabnoteb{$\pm$ 0.1} & 37.1 \tabnoteb{$\pm$ 0.0} & 37.1 \tabnoteb{$\pm$ 0.1} &  
37.1
\\ \midrule \midrule

\multirow{2}{*}{\textbf{\textit{S}\ =\ 20}} & Raw & \textit{R}\ =\ 1 & \textit{R}\ =\ 2 &
\textit{R}\ =\ 3 & \textit{R}\ =\ 4 & \textit{R}\ =\ 5 & \textit{R}\ =\ 6 & \textit{R}\ =\ 7 & \textit{R}\ =\ 8 & \textit{R}\ =\ 9 & \textit{R}\ =\ 10 \\ \cmidrule(l){2-12} &

15.47 & 17.6 \tabnoteb{$\pm$ 1.2} & 17.5 \tabnoteb{$\pm$ 0.7} & 17.6 \tabnoteb{$\pm$ 0.7} & 17.8 \tabnoteb{$\pm$ 0.4} & 17.5 \tabnoteb{$\pm$ 0.3} & 17.7 \tabnoteb{$\pm$ 0.4} & 17.8 \tabnoteb{$\pm$ 0.3} & 17.6 \tabnoteb{$\pm$ 0.2} & 17.7 \tabnoteb{$\pm$ 0.1} & 17.7
\\ \bottomrule
    \end{tabular}
    }

    \end{center}
    \vspace{-4mm}
    \caption{Average top-1 incremental accuracy of \adbic-corrected models trained incrementally on \food with \lucir, for $S=\{5, 10, 20\}$ states, while varying the number $R$ of reference datasets. For $R \leq 9$, results are averaged across 10 random samplings of the reference datasets (hence the std values). \textit{Raw} is the accuracy of \lucir without bias correction.}
    \vspace{-4mm}
    \label{tab:ablation_datasets}
\end{table*}

\subsection{Robustness of dataset knowledge transfer}

We complement the results presented in Table~\ref{tab:global} with two experiments which further evaluate the robustness of \adbic. 
First, we test the effect of a different number of training images per class for reference and target datasets.
We remove $50\%$ of training images for target datasets to test the transferability in this setting.
The obtained results, presented in Table~\ref{tab:ablation_train}, indicate that performance gains are systematic for \lwf and \lucir, albeit lower compared to results in Table~\ref{tab:global}.
Results are more mixed for \siw and \ftplus, but \adbic still has a positive effect in the majority of cases. 
This experiment shows that the proposed dataset knowledge transfer approach is usable for reference and target datasets which have a different number of training samples per class.
However, maintaining a low difference in dataset sizes is preferable in order to keep the transfer effective.

Second, we assess the robustness of the method with respect to $R$, the number of available reference datasets. We select the \food dataset because it has the largest domain shift with respect to reference datasets and is thus the most suitable for this experiment.
We vary $R$ from 1 to 9 and perform transfer with ten random samplings for each $R$ value.
Results obtained for \lucir are reported in Table~\ref{tab:ablation_datasets}.
Accuracy levels are remarkably stable for different values of $R$ and significant gains are obtained even when using a single reference dataset. 
These results confirm that parameter transfer is effective even with few reference datasets, which is interesting considering that the computational cost of offline training is also reduced.
Results with other methods for \cifar are provided in the supp. material.

\section{Conclusion}
We introduced a method which enables the use of bias correction methods for class IL without memory.
This IL scenario is challenging, because catastrophic forgetting is very strong in the absence of memory. 
The proposed method \textit{TransIL} transfers bias correction parameters learned offline from reference datasets toward target datasets.
Since reference dataset training is done offline, a validation memory which includes exemplars from all incremental states can be exploited to optimize the bias correction layer.
The evaluation provides comprehensive empirical support for the transferability of bias correction parameters.
Performance is improved for all but two of the configurations tested, with gains up to 16 top-1 accuracy points.
Robustness evaluation shows that parameter transfer is efficient when only a small number of reference datasets is used for transfer. 
It is also usable when the number of training images per class in target datasets is different from that of available reference datasets.
These last two findings are important in practice since the same reference datasets can be exploited in different incremental configurations.
A second contribution relates to the modeling of the degree of forgetting associated to past states.
While recency bias was already acknowledged~\cite{masana2021_study}, no difference was made between past classes learned in different IL states~\cite{wu2019_bic}.
This is in part due to validation memory constraints which appear when the bias correction layer is optimized during the incremental process. 
Such constraints are reduced here since reference datasets training is done offline and a refined definition of the \bic layer with specific parameters for each past state becomes possible.
The comparison of the standard and of the proposed definition of the bias correction layer is favorable to the latter.
The reported results encourage us to pursue the work presented here.
First, the parameter transfer is done using average values of parameters learned on reference datasets.
A finer-grained transfer method will be tested to get closer to the oracle results reported in Table~\ref{tab:global}.
The objective is to automatically select the best reference dataset in each IL state of a target dataset.
Second, we exploit an adapted version of a bias correction method which was initially designed for class IL with memory.
We will explore the design of methods which are specifically created for class IL without memory. 
Finally, while distillation-based methods outperformed methods which do not use distillation for the datasets tested here, existing results~\cite{belouadah2020_siw,masana2021_study} indicate that the role of distillation diminishes with scale.
It would be interesting to verify this finding for our method.

\vspace{2mm}

\noindent \textbf{Acknowledgments.}
This work was supported by the European Commission under European Horizon 2020 Programme, grant number 951911 - AI4Media, and partially supported by the LabEx PERSYVAL-Lab (ANR-11-LABX-0025-01) funded by the French program "Investissements d'Avenir". This publication was made possible by the use of the FactoryIA supercomputer, financially supported by the Ile-de-France Regional Council.

{\small
\bibliographystyle{ieee_fullname}
\bibliography{egbib}
}

\newpage

\begin{minipage}{\textwidth}
    \vskip .375in
    \begin{center}
      {\Large \bf Supplementary Material for "Dataset Knowledge Transfer for Class-Incremental Learning without Memory" \par}
      \vspace*{24pt}
      {
      \large
      \lineskip .5em
      \begin{tabular}[t]{c}
            Habib Slim$^{1*}$\hspace{1em}
            Eden Belouadah$^{1, 2*}$\hspace{1em}
            Adrian Popescu$^{1}$\hspace{1em}
            Darian Onchis$^{3}$ \\
            $^{1}$Université Paris-Saclay, CEA, List, F-91120, Palaiseau, France\\
            $^{2}$IMT Atlantique, Lab-STICC, team RAMBO, UMR CNRS 6285, F-29328, Brest, France\\
            $^{3}$ West University of Timisoara, Timisoara, Romania \\
            {\tt\footnotesize habib.slim@grenoble-inp.org, \{eden.belouadah, adrian.popescu\}@cea.fr, darian.onchis@e-uvt.ro}
         \vspace*{1pt}\\
      \end{tabular}
      \par
      }

      \vskip .5em
      \vspace*{12pt}
    \end{center}
\end{minipage}
\thispagestyle{empty}


\appendix
\section*{Introduction}
\begin{figure*}[bht]
    \centering
    
    \caption*{\large \hspace{1.5em}\cifar\hspace{16.5em}\imnet\hspace{1em}}

    \subfloat{\includegraphics[width=0.46\textwidth]{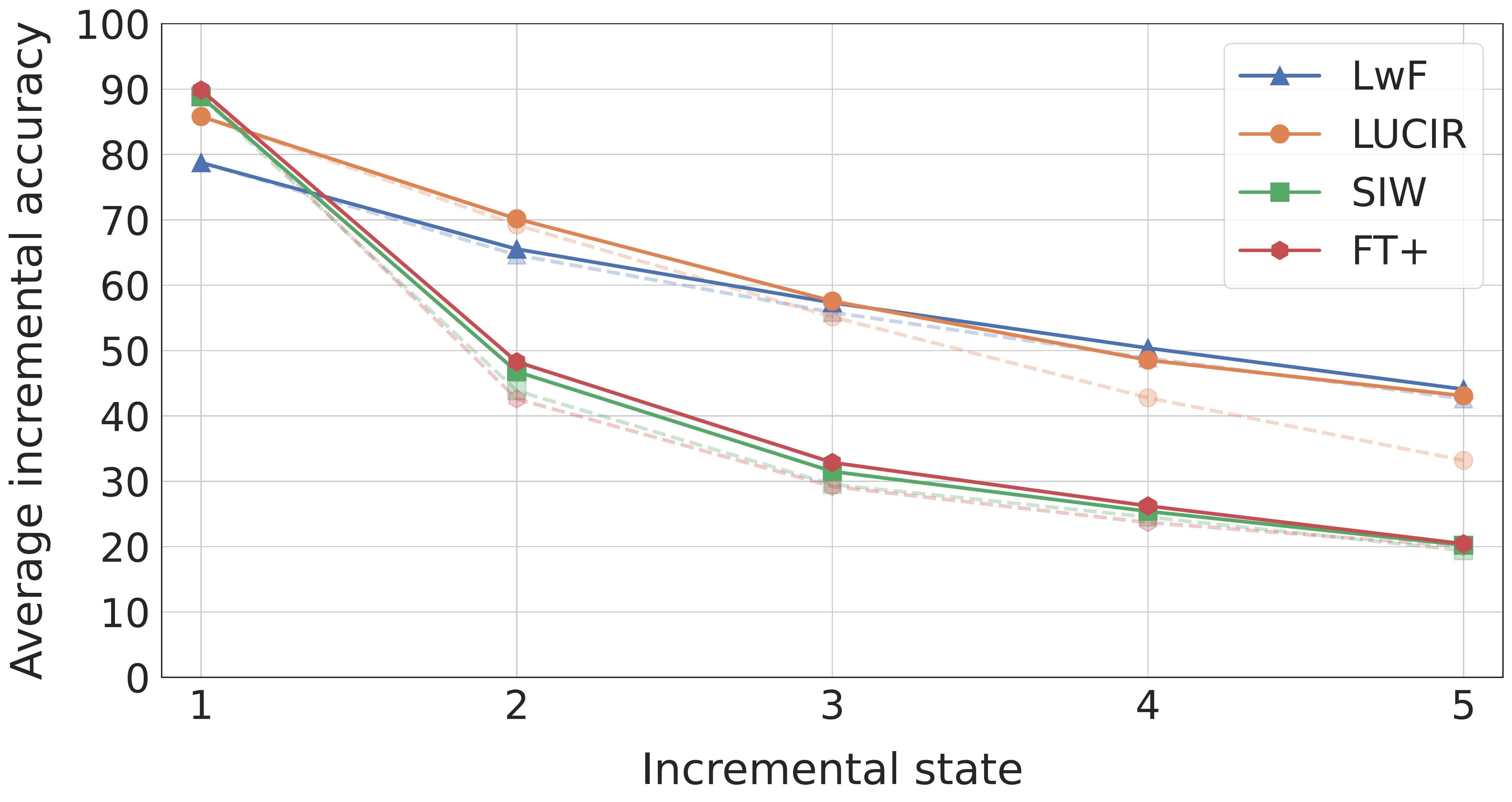}}
    \hspace{2em}
    \subfloat{\includegraphics[width=0.46\textwidth]{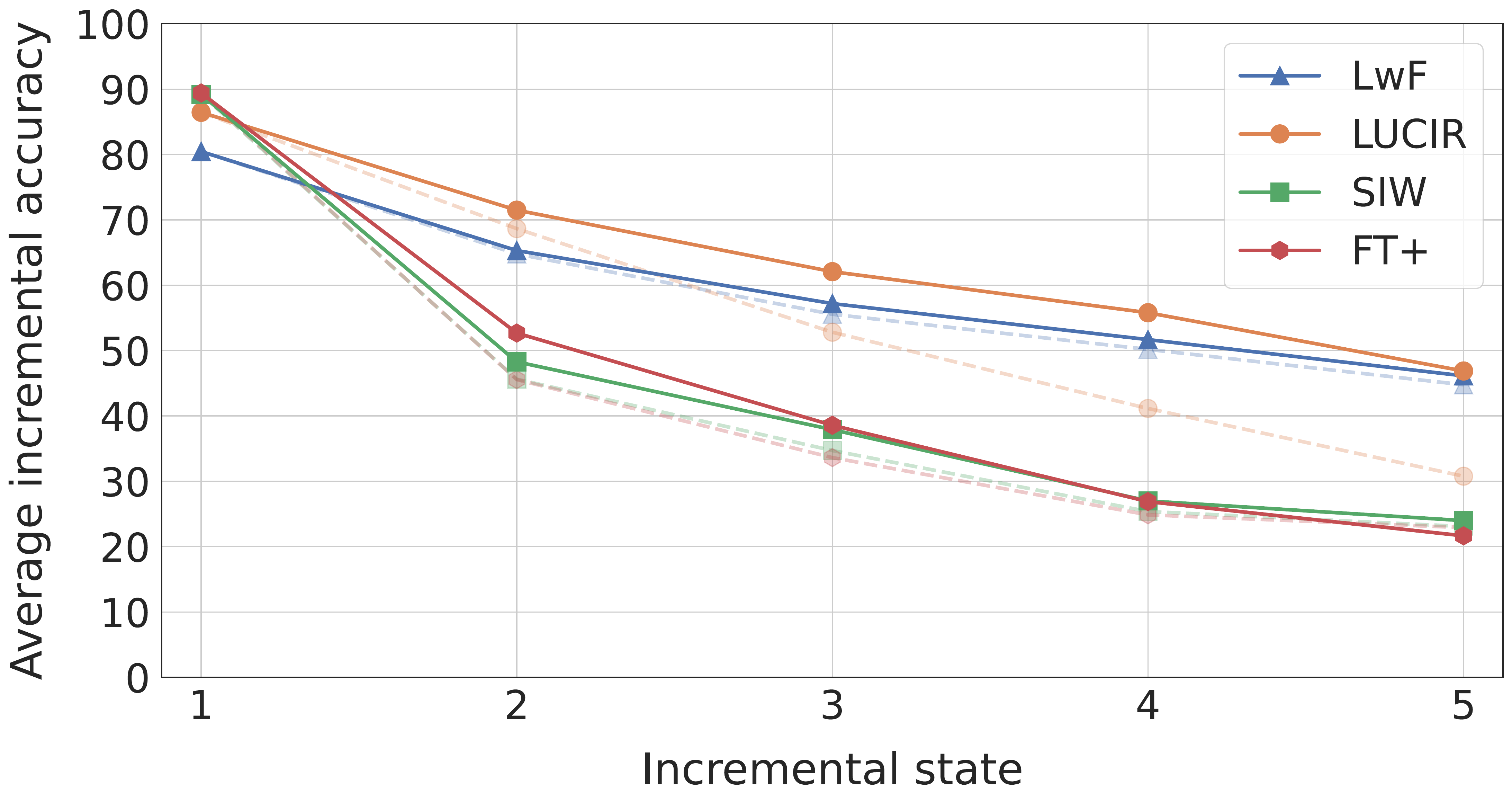}}
    
    \vspace{1em}
    
    \subfloat{\includegraphics[width=0.46\textwidth]{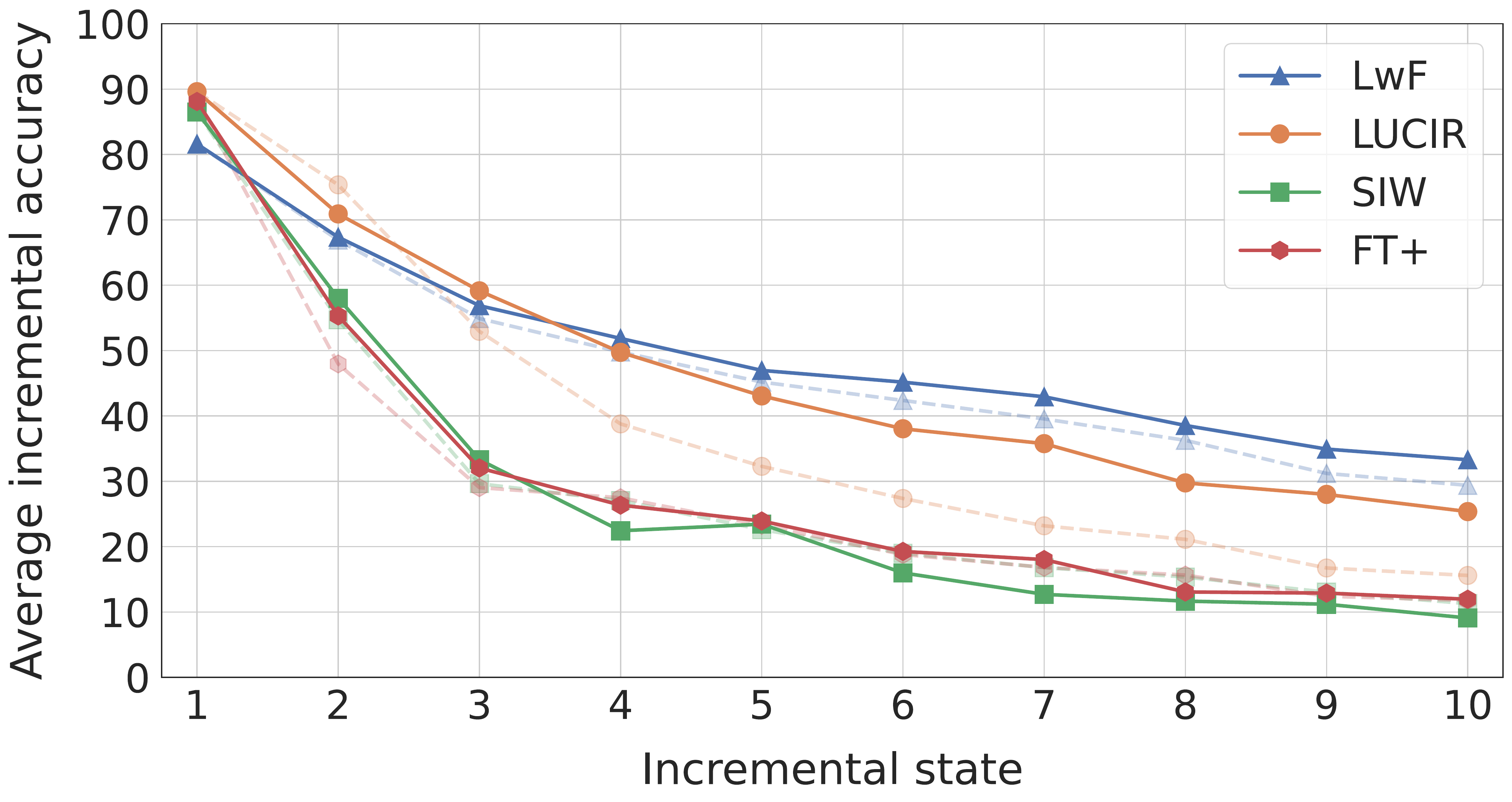}}
    \hspace{2em}
    \subfloat{\includegraphics[width=0.46\textwidth]{images_supp_material/full_plots/all_plot_cif100_s10.pdf}}
    
    \vspace{1em}
    
    \subfloat{\includegraphics[width=0.46\textwidth]{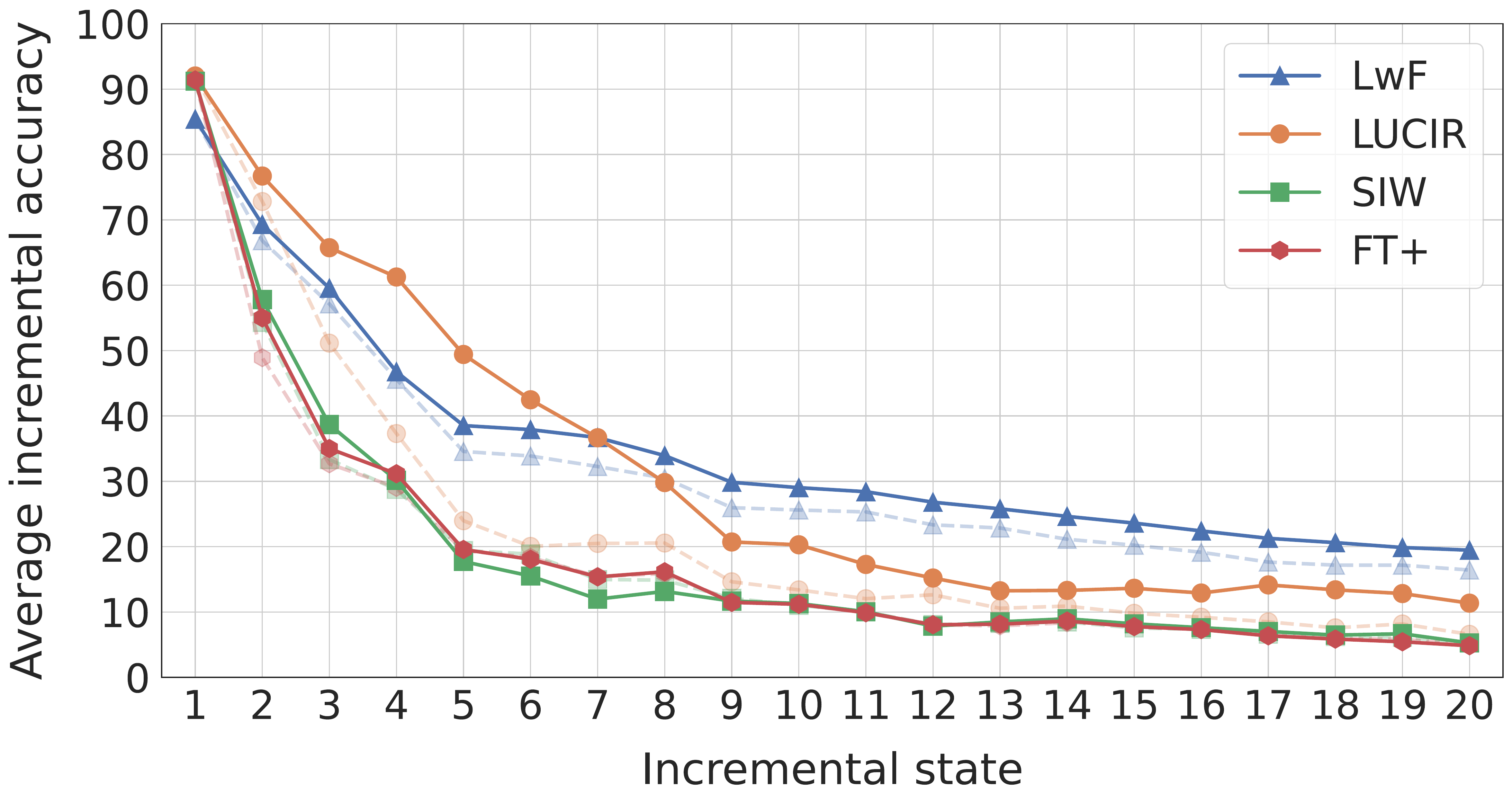}}
    \hspace{2em}
    \subfloat{\includegraphics[width=0.46\textwidth]{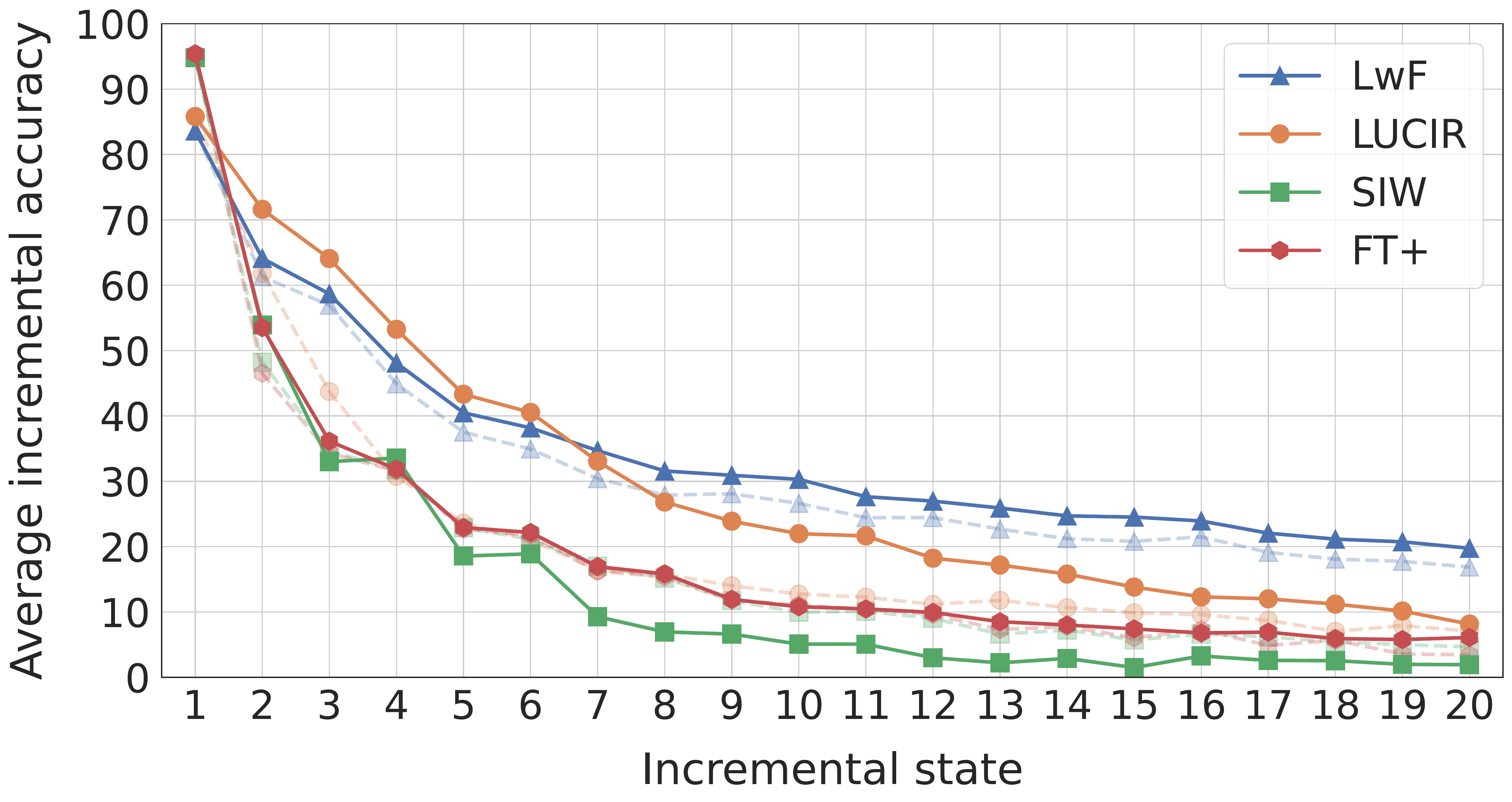}}

    \vspace{2em}
        
    \caption[Detailed $TransIL$ results on \cifar and \imnet]{Average accuracies in each state on \cifar (\textit{left}) and \imnet (\textit{right}) datasets with all backbone methods after \adbic correction, for $S=5$ (\textit{top}), $S=10$ (\textit{middle}) and $S=20$ (\textit{bottom}) states. The accuracies without correction of the corresponding methods are provided in dashed lines (same colors). \textit{Best viewed in color}.}
	\label{fig:full_plots1}
\end{figure*}

\begin{figure*}[bht]
    \centering
    
    \caption*{\large \hspace{1.75em}\birds\hspace{16.5em}\food\hspace{1em}}

    \subfloat{\includegraphics[width=0.46\textwidth]{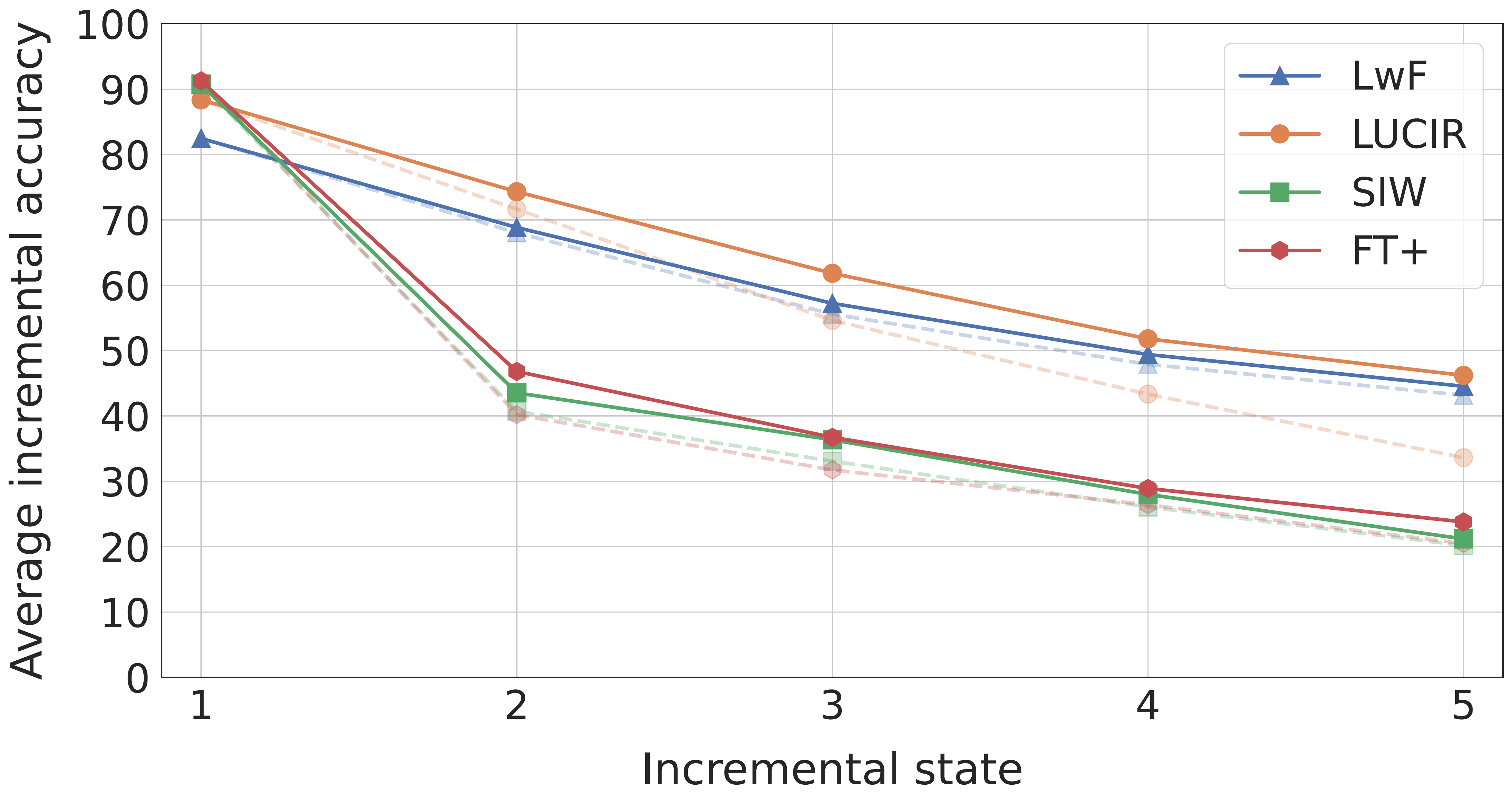}}
    \hspace{2em}
    \subfloat{\includegraphics[width=0.46\textwidth]{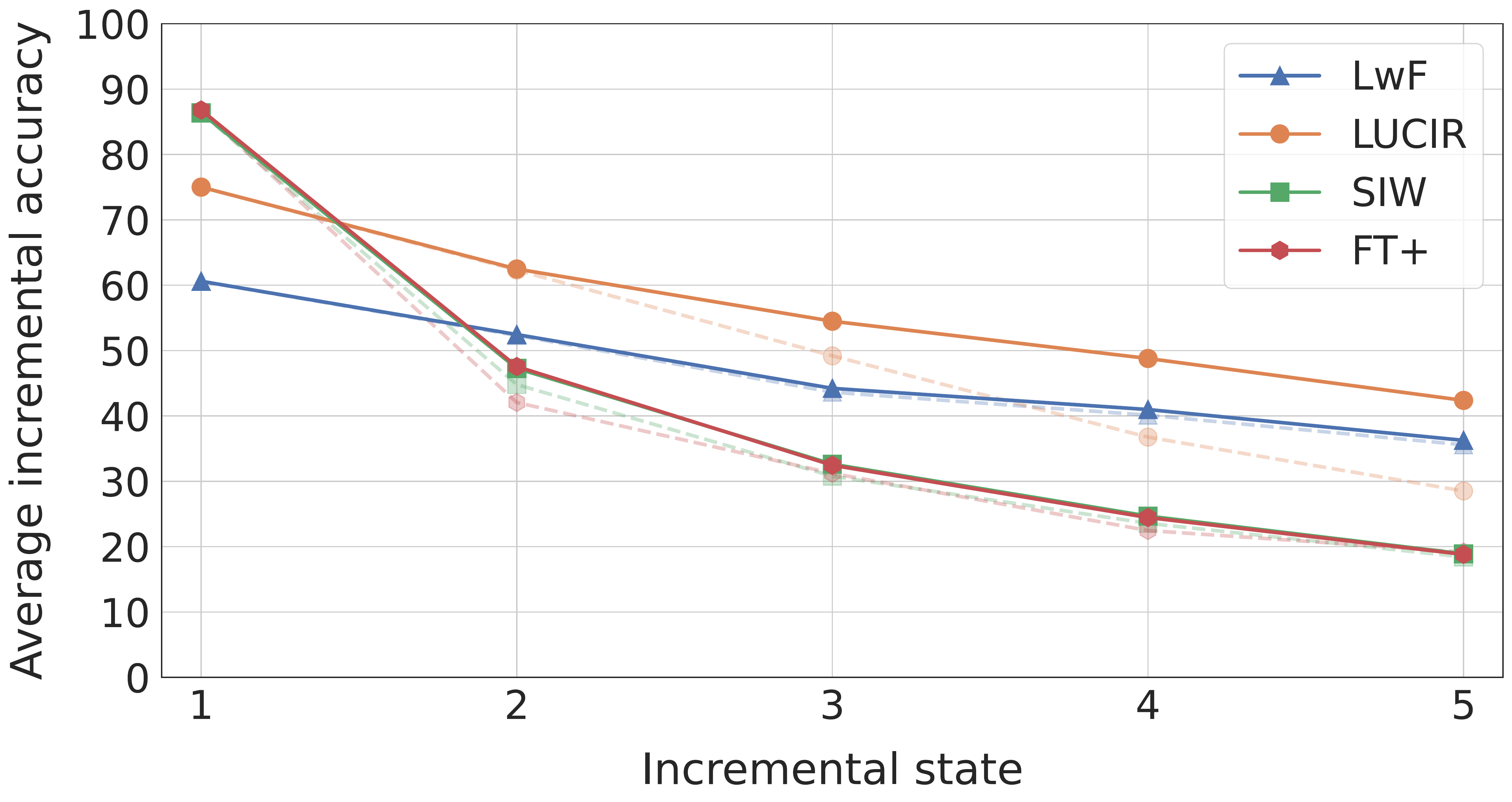}}
    
    \vspace{1em}
    
    \subfloat{\includegraphics[width=0.46\textwidth]{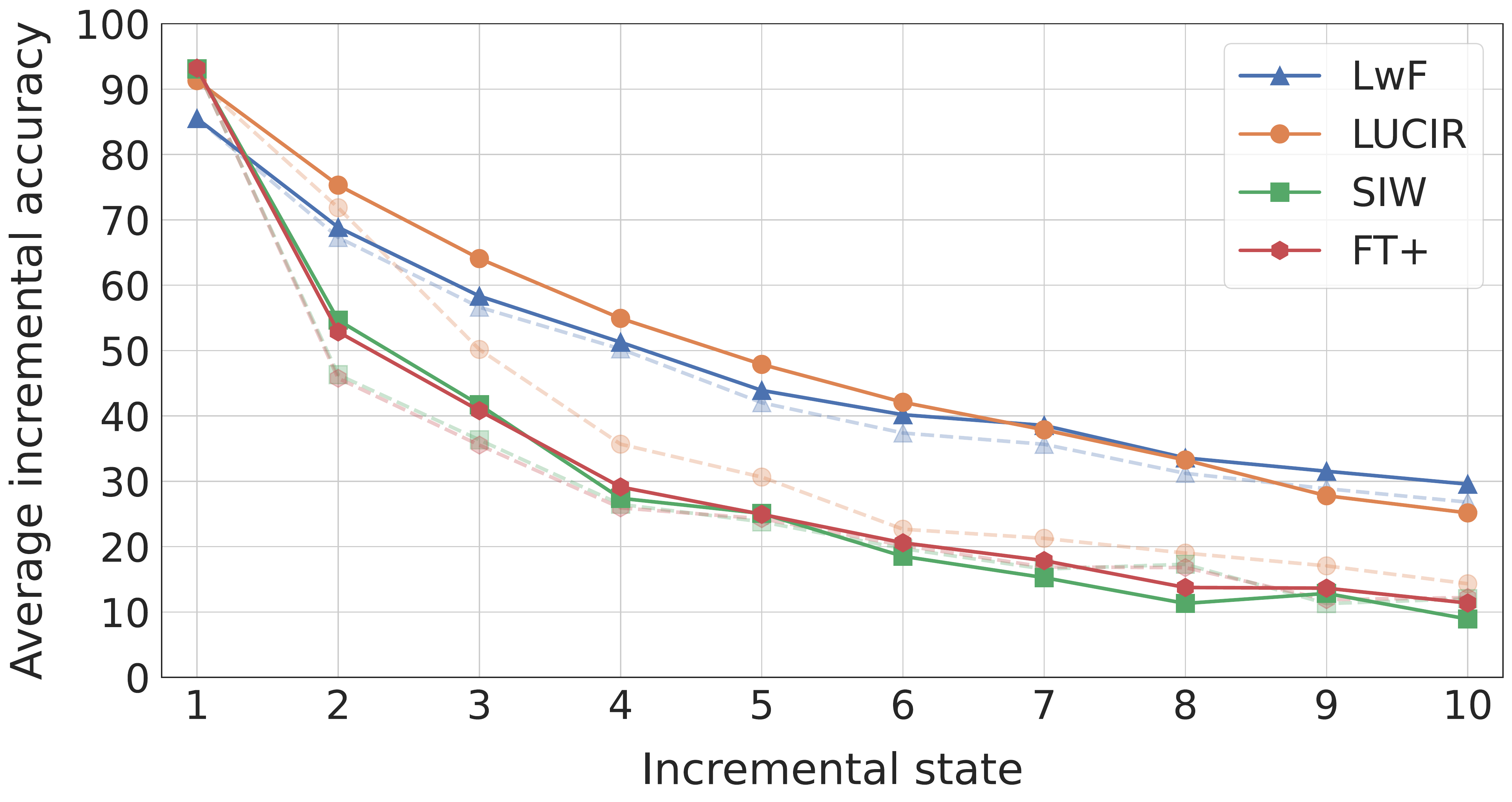}}
    \hspace{2em}
    \subfloat{\includegraphics[width=0.46\textwidth]{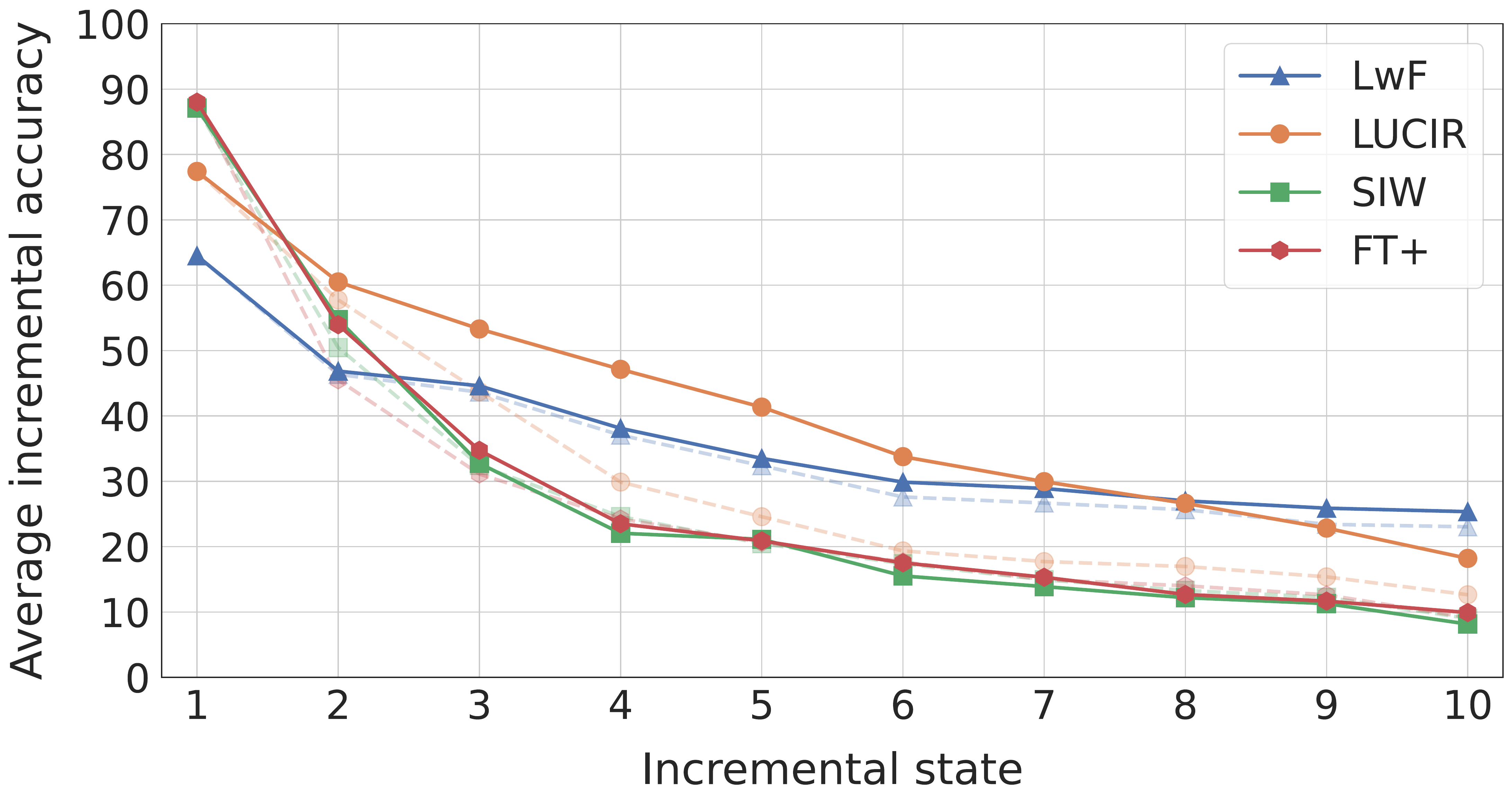}}
    
    \vspace{1em}
    
    \subfloat{\includegraphics[width=0.46\textwidth]{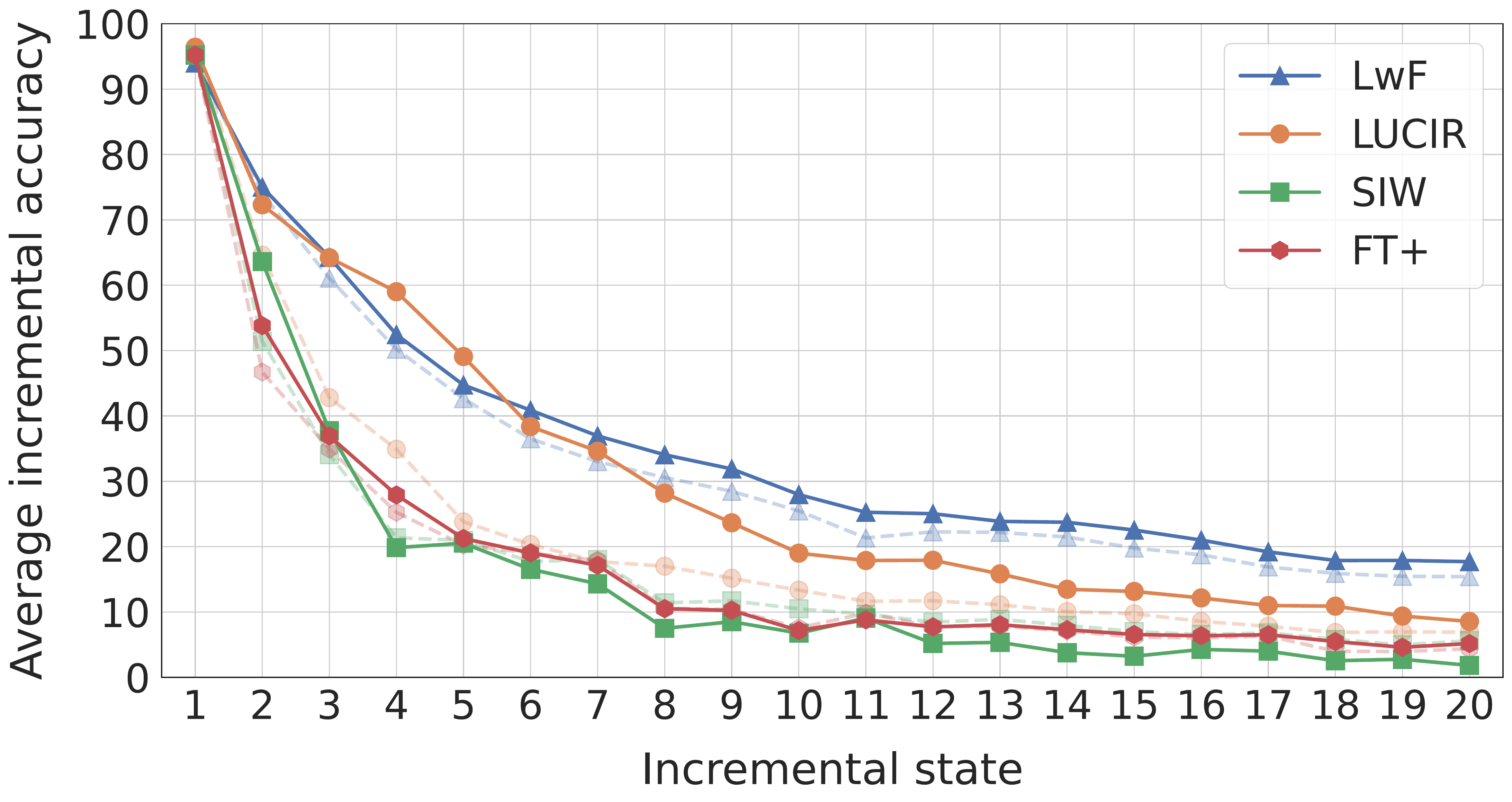}}
    \hspace{2em}
    \subfloat{\includegraphics[width=0.46\textwidth]{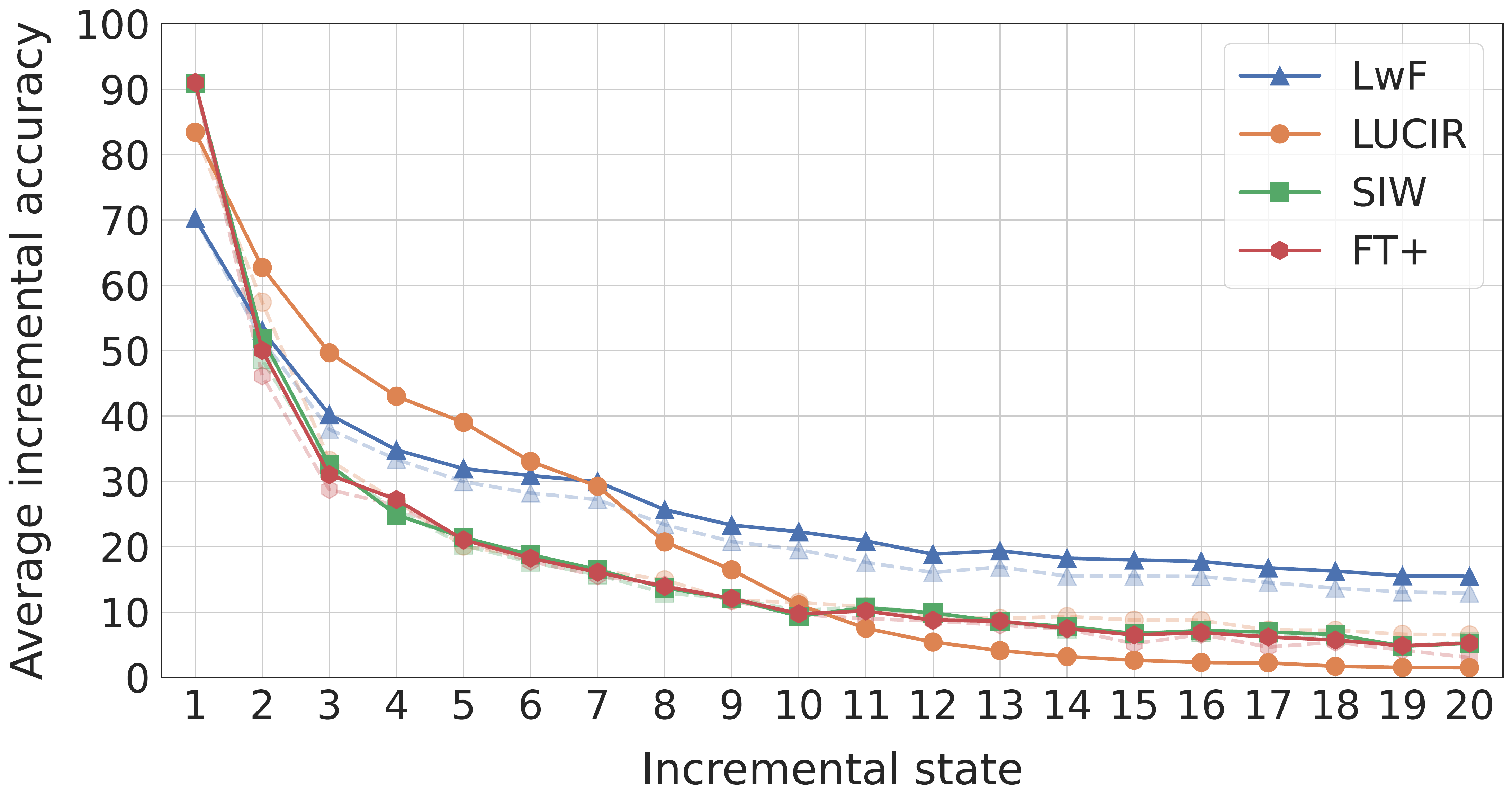}}

    \vspace{2em}
        
    \caption[Detailed $TransIL$ results on \birds and \food]{Average accuracies in each state on \birds (left) and \food (right) datasets with all backbone methods after \adbic correction, for $S=5$ (\textit{top}), $S=10$ (\textit{middle}) and $S=20$ (\textit{bottom}) states. The accuracies without correction of the corresponding methods are provided in dashed lines (same colors). \textit{Best viewed in color}.}
	\label{fig:full_plots2}
\end{figure*}

\noindent \ In this supplementary material, we provide:

\vspace{-0.5em}
\begin{itemize}
    \setlength\itemsep{0.1em}
    \item implementation details of \adbic and the tested backbone IL methods (Section \ref{sec:imp_details}).
    \item classes lists of target datasets used for evaluation (Section \ref{sec:datasets}).
    \item additional figures highlighting the effects of \adbic on the tested backbone methods (Section \ref{sec:state_wise_accs}).
    \item additional tables for the robustness experiment presented in Section 4.4 of the paper (Section \ref{sec:varying_nb_datasets}).
    \item results on \places dataset (Table \ref{tab:places365}).
    \item additional accuracy plots for all methods and datasets (Figures \ref{fig:full_plots1} and \ref{fig:full_plots2}).
\end{itemize}

\begin{figure*}[bht]
    \centering
    \captionsetup[subfigure]{justification=justified, labelformat=empty, width=150em}

    \begin{subfigure}{\textwidth}
        \centering
        \subfloat{\includegraphics[width=\textwidth]{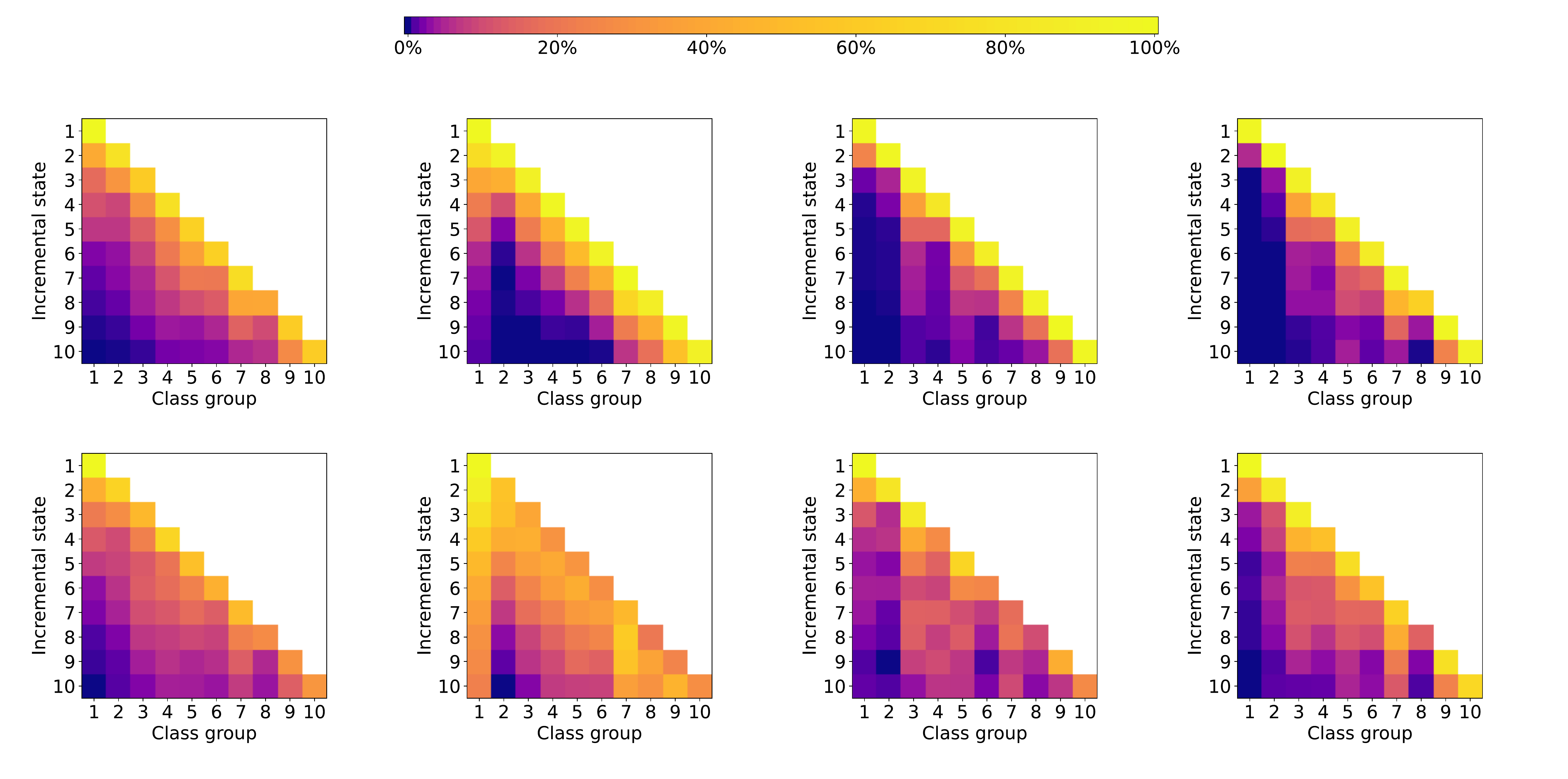}}
    \end{subfigure}
    
    \begin{subfigure}{0.92\textwidth}
        \centering
        \begin{multicols}{4}
        (a)\ \ LwF \cite{li2016_lwf}
        \columnbreak
        
        (b)\ \ LUCIR \cite{hou2019_lucir}
        \columnbreak
        
        (c)\ \ SIW \cite{belouadah2020_siw}
        \columnbreak
        
        (d)\ \ FT+ \cite{masana2021_study}
        \end{multicols}
    \end{subfigure}
    \vspace{0.25em}

    \caption{Accuracies per incremental state for each class group, for models trained with \lwf and \lucir on \cifar for $S=10$ states, before (\textit{top}) and after (\textit{bottom}) \adbic correction. Each row represents an incremental state and each square the accuracy on a group of classes first learned in a specific state. In the first state, represented by the first rows of the matrices, models are only evaluated on the first class group. In the second state, represented by the second rows, models are evaluated on the first two class groups, etc. \textit{Best viewed in color}.}
	\label{fig:state_wise_acc}
    \vspace{2em}
\end{figure*}

\vspace{-1em}
\section{Implementation details}
\label{sec:imp_details}

\subsection{Backbone IL methods}

For \lucir~\cite{hou2019_lucir} and \siw~\cite{belouadah2020_siw}, we used the original codes provided by the authors. For \lwf, we adapted the multi-class Tensorflow \cite{abadi2015_tensorflow} implementation from \cite{rebuffi2017_icarl} to IL without memory. For \ftplus, we implemented the method by replacing classification weights of each class group by their initial weights learned when classes were encountered for the first time.\\

All methods use a ResNet-18 \cite{he2016_resnet} backbone, with batch size $128$. For \lwf, we use a base learning rate of $1.0$ divided by $5$ after $20$, $30$, $40$ and $50$ epochs. The weight decay is set to $10^{-5}$ and models are trained for $70$ epochs in each state. For \lucir, we mostly use the parameters recommended for \cifar in the original paper \cite{hou2019_lucir}. We set $\lambda_{\text{base}}$ to $5$. For each state, we train models for $160$ epochs. The base learning rate is set to $0.1$ and divided by $10$ after $80$ and $120$ epochs.\\ \vfill\null

\vspace{22em}
The weight decay is set to $5 \cdot 10^{-4}$ and the momentum to $0.9$. Note that since no memory of past classes is available, the margin ranking loss is unusable and thus removed. \siw and \ftplus are both trained with the same set of hyperparameters. Following \cite{belouadah2020_siw}, models are trained from scratch for $300$ epochs in the first non-incremental state, using the SGD optimizer with momentum $0.9$.\\

The base learning rate is set to $0.1$, and is divided by $10$ when the loss plateaus for $60$ epochs. The weight decay is set to $5 \cdot 10^{-4}$. 
For incremental states, the same hyperparameters are used, except for the number of epochs which is reduced to $70$ and for the learning rate which is divided by $10$ when the loss plateaus for $15$ epochs.

\begin{table*}[th]
\small
    \centering
    \begin{tabularx}{\linewidth}{c|X}
    \toprule
    & \multicolumn{1}{c}{\textbf{Classes names}}  \\
    \midrule
    \cifar & Apple, Aquarium fish, Baby, Bear, Beaver, Bed, Bee, Beetle, Bicycle, Bottle, Bowl, Boy, Bridge, Bus, Butterfly, Camel, Can, Castle, Caterpillar, Cattle, Chair, Chimpanzee, Clock, Cloud, Cockroach, Couch, Crab, Crocodile, Cup, Dinosaur, Dolphin, Elephant, Flatfish, Forest, Fox, Girl, Hamster, House, Kangaroo, Keyboard, Lamp, Lawn mower, Leopard, Lion, Lizard, Lobster, Man, Maple tree, Motorcycle, Mountain, Mouse, Mushroom, Oak tree, Orange, Orchid, Otter, Palm tree, Pear, Pickup truck, Pine tree, Plain, Plate, Poppy, Porcupine, Possum, Rabbit, Raccoon, Ray, Road, Rocket, Rose, Sea, Seal, Shark, Shrew, Skunk, Skyscraper, Snail, Snake, Spider, Squirrel, Streetcar, Sunflower, Sweet pepper, Table, Tank, Telephone, Television, Tiger, Tractor, Train, Trout, Tulip, Turtle, Wardrobe, Whale, Willow tree, Wolf, Woman, Worm \\ \hline
    
    \imnet & Bletilla striata, Christmas stocking, Cognac, European sandpiper, European turkey oak, Friesian, Japanese deer, Luger, Sitka spruce, Tennessee walker, Torrey pine, Baguet, Bald eagle, Barn owl, Bass guitar, Bathrobe, Batting helmet, Bee eater, Blue gum, Blue whale, Bones, Borage, Brass, Caftan, Candytuft, Carthorse, Cattle egret, Cayenne, Chairlift, Chicory, Cliff dwelling, Cocktail dress, Commuter, Concert grand, Crazy quilt, Delivery truck, Detached house, Dispensary, Drawing room, Dress hat, Drone, Frigate bird, Frozen custard, Gemsbok, Giant kangaroo, Guava, Hamburger bun, Hawfinch, Hill myna, Howler monkey, Huisache, Jennet, Jodhpurs, Ladder truck, Loaner, Micrometer, Mink, Moorhen, Moorhen, Moped, Mortarboard, Mosquito net, Mountain zebra, Muffler, Musk ox, Obelisk, Opera, Ostrich, Ox, Oximeter, Playpen, Post oak, Purple-fringed orchid, Purple willow, Quaking aspen, Ragged robin, Raven, Redpoll, Repository, Roll-on, Scatter rug, Shopping cart, Shower curtain, Slip-on, Spider orchid, Sports car, Steam iron, Stole, Stuffed mushroom, Subcompact, Sundial, Tabby, Tabi, Tank car, Tramway, Unicycle, Wagtail, Walker, Window frame, Wood anemone \\  \hline
   
    \birds & American bittern, American coot, Atlantic puffin, Baltimore oriole, Barrow's goldeneye, Bewick's swan, Bullock's oriole, California quail, Eurasian kingfisher, European gallinule, European sandpiper, Orpington, Amazon, Barn owl, Black-crowned night heron, Black-necked stilt, Black-winged stilt, Black swan, Black vulture, Black vulture, Blue peafowl, Brambling, Bufflehead, Buzzard, Cassowary, Cockerel, Common spoonbill, Crossbill, Duckling, Eastern kingbird, Emperor penguin, Fairy bluebird, Fishing eagle, Fulmar, Gamecock, Golden pheasant, Goosander, Goshawk, Great crested grebe, Great horned owl, Great white heron, Greater yellowlegs, Greenshank, Gyrfalcon, Hawfinch, Hedge sparrow, Hen, Honey buzzard, Hornbill, Kestrel, Kookaburra, Lapwing, Least sandpiper, Little blue heron, Little egret, Macaw, Mandarin duck, Marsh hawk, Moorhen, Mourning dove, Muscovy duck, Mute swan, Ostrich, Owlet, Oystercatcher, Pochard, Raven, Red-legged partridge, Red-winged blackbird, Robin, Robin, Rock hopper, Roseate spoonbill, Ruby-crowned kinglet, Ruffed grouse, Sanderling, Screech owl, Screech owl, Sedge warbler, Shoveler, Siskin, Snow goose, Snowy egret, Song thrush, Spotted flycatcher, Spotted owl, Sulphur-crested cockatoo, Thrush nightingale, Tropic bird, Tufted puffin, Turkey cock, Weka, Whistling swan, White-breasted nuthatch, White-crowned sparrow, White-throated sparrow, White stork, Whole snipe, Wood ibis, Wood pigeon.  \\  \hline
    
    \food & Apple pie, Baby back ribs, Baklava, Beef carpaccio, Beef tartare, Beet salad, Beignets, Bibimbap, Bread pudding, Breakfast burrito, Bruschetta, Caesar salad, Cannoli, Caprese salad, Carrot cake, Ceviche, Cheese plate, Cheesecake, Chicken curry, Chicken quesadilla, Chicken wings, Chocolate cake, Chocolate mousse, Churros, Clam chowder, Club sandwich, Crab cakes, Creme brulee, Croque madame, Cup cakes, Deviled eggs, Donuts, Dumplings, Edamame, Eggs benedict, Escargots, Falafel, Filet mignon, Fish and chips, Foie gras, French fries, French onion soup, French toast, Fried calamari, Fried rice, Frozen yogurt, Garlic bread, Gnocchi, Greek salad, Grilled cheese sandwich, Grilled salmon, Guacamole, Gyoza, Hamburger, Hot and sour soup, Hot dog, Huevos rancheros, Hummus, Ice cream, Lasagna, Lobster bisque, Lobster roll sandwich, Macaroni and cheese, Macarons, Miso soup, Mussels, Nachos, Omelette, Onion rings, Oysters, Pad thai, Paella, Pancakes, Panna cotta, Peking duck, Pho, Pizza, Pork chop, Poutine, Prime rib, Pulled pork sandwich, Ramen, Ravioli, Red velvet cake, Risotto, Samosa, Sashimi, Scallops, Seaweed salad, Shrimp and grits, Spaghetti bolognese, Spaghetti carbonara, Spring rolls, Steak, Strawberry shortcake, Sushi, Tacos, Takoyaki, Tiramisu, Tuna tartare  \\  
    \bottomrule
    \end{tabularx}
\end{table*}

\begin{table*}[th]
    \small
    \centering
    \begin{tabularx}{\linewidth}{c|X}
    \toprule
    & \multicolumn{1}{c}{\textbf{Classes names}}  \\
    \midrule

    \places & Airplane cabin, Amphitheater, Amusement arcade, Aqueduct, Arcade, Archaelogical excavation, Archive, Arena performance, Attic, Bamboo forest, Bar, Barn, Baseball field, Bazaar outdoor, Beach, Beach house, Beauty salon, Bedroom, Bookstore, Bus interior, Cafeteria, Castle, Chemistry lab, Church outdoor, Cliff, Corridor, Courthouse, Crevasse, Department store, Desert sand, Desert vegetation, Dining room, Dorm room, Drugstore, Elevator lobby, Elevator shaft, Entrance hall, Escalator indoor, Farm, Field cultivated, Field wild, Florist shop indoor, Food court, Fountain, Garage indoor, Gazebo exterior, Golf course, Hangar outdoor, Harbor, Hardware store, Hayfield, Heliport, Highway, Home theater, Hospital room, Hot spring, Hotel outdoor, Hunting lodge outdoor, Ice skating rink indoor, Junkyard, Kasbah, Kitchen, Lagoon, Lake natural, Marsh, Mosque outdoor, Oast house, Office cubicles, Pagoda, Park, Pavilion, Physics laboratory, Pier, Porch, Racecourse, Residential neighborhood, Restaurant, Rice paddy, Rock arch, Ruin, Sauna, Server room, Shed, Shopfront, Storage room, Sushi bar, Television room, Television studio, Throne room, Topiary garden, Tower, Tree house, Trench, Underwater ocean deep, Waiting room, Water park, Waterfall, Wet bar, Windmill, Zen garden\\
    \bottomrule
    \end{tabularx}
    \caption{Classes names of target datasets listed by alphabetical order}
    \label{tab:classes_names}
\end{table*}

\begin{table*}[htb!]
    \begin{center}
    \begin{tabular}{@{\kern0.5em}lccccccc@{\kern0.5em}}
        \toprule
        \multirow{2}{*}{\textbf{Method}}
            & \multicolumn{3}{c}{\places}
            && \multicolumn{3}{c}{\places\ \ (\textit{halved})}
        \\ \cmidrule(lr){2-4} \cmidrule(lr){6-8}
        
            & \multicolumn{1}{c}{\textit{S}\ =\ 5}
            & \multicolumn{1}{c}{\textit{S}\ =\ 10}
            & \multicolumn{1}{c}{\textit{S}\ =\ 20}
            &
            & \multicolumn{1}{c}{\textit{S}\ =\ 5}
            & \multicolumn{1}{c}{\textit{S}\ =\ 10}
            & \multicolumn{1}{c}{\textit{S}\ =\ 20}
        \\ \midrule \midrule

        \addlinespace[0.25em]
        \textbf{LwF} \cite{li2016_lwf}                           
            & 43.3 & 35.1 & 25.9 && 35.4 & 27.7 & 21.5 \\

            \rowcolor[RGB]{240,240,240}
            \textit{w}/ \adbic    
            &  \textbf{44.2} \tabnoteg{+ 0.9} & \textbf{36.6} \tabnoteg{+ 1.5} & \textbf{28.6} \tabnoteg{+ 2.7} 
            && 35.9 \tabnoteg{+ 0.5} & 28.5 \tabnoteg{+ 0.8} & \textbf{23.6} \tabnoteg{+ 2.1} \\
        \midrule
        \addlinespace[0.25em]

        \textbf{LUCIR} \cite{hou2019_lucir}
            & 40.5 & 26.0 & 16.0 && 35.5 & 23.2 & 14.7 \\

            \rowcolor[RGB]{240,240,240}
            \textit{w}/ \adbic    
            &  42.8 \tabnoteg{+ 2.3} & 35.4 \tabnoteg{+ 9.4} & 23.3 \tabnoteg{+ 7.3}
            && \textbf{40.5} \tabnoteg{+ 5.0} & \textbf{33.6} \tabnoteg{+ 10.4} & 22.3 \tabnoteg{+ 7.6} \\
        \midrule
        \addlinespace[0.25em]

        \textbf{SIW} \cite{belouadah2020_siw}                           
            & 27.3 & 20.6 & 14.0 && 27.2 & 19.6 & 14.8 \\

            \rowcolor[RGB]{240,240,240}
            \textit{w}/ \adbic
            &  28.8 \tabnoteg{+ 1.5} & 21.2 \tabnoteg{+ 0.6} & 13.1 \tabnoter{- 0.9}
            && 28.5 \tabnoteg{+ 1.3} & 19.3 \tabnoter{- 0.3} & 14.3 \tabnoter{- 0.5} \\
        \midrule
        \addlinespace[0.25em]

        \textbf{FT+} \cite{masana2021_study}            
            & 26.9 & 20.8 & 12.1 && 00.0 & 00.0 & 00.0 \\

            \rowcolor[RGB]{240,240,240}
            \textit{w}/ \adbic    
            &  27.3 \tabnoteg{+ 0.4} & 19.7 \tabnoter{- 1.1} & 13.2 \tabnoteg{+ 1.1}
            && 25.6 \tabnoter{- 0.5} & 17.2 \tabnoter{- 2.7} & 13.5 \tabnoteg{+ 1.1} \\
        \bottomrule
        \end{tabular}
    \end{center}
    \caption{Average top-1 incremental accuracy using $S=\{5,10,20\}$ states, for the \places dataset with all and half of the training data. The \places dataset is extracted from Places365 \cite{zhou2017_places}. Similarly to the other datasets presented in the main paper, we randomly select a hundred classes from the original dataset to construct a dataset with a hundred classes. Results obtained are comparable to those obtained on the other datasets, despite the domain shift from ImageNet. Gains obtained over the backbone method are given in green, and the best results for each setting in bold.}
    \label{tab:places365}
\end{table*}

\subsection{Adaptive bias correction}
The correction of output scores is done in the same way for all methods. After the extraction of scores and labels for each model, batches are fed into a PyTorch \cite{paszke2019_pytorch} module which performs the optimization of \adbic parameters, or the transfer of previously learned parameters.
Following \cite{masana2021_study}, \bic and \adbic layers are implemented as pairs of parameters and optimized simply through backpropagation.\\

Parameters $\alpha_s^k, \beta_s^k$ corresponding to each incremental state $s$ are optimized for 300 epochs, with the Adam \cite{kingma2015_adam} optimizer and a starting learning rate of $10^{-3}$. An L2-penalty is added to the loss given in Equation \ref{eq:adaptive_bic_optim} of the main paper, with a lambda of $5 \cdot 10^{-3}$ for $\alpha$ parameters and $5 \cdot 10^{-2}$ for $\beta$ parameters.

\newpage
\section{Datasets description}
\label{sec:datasets}

We provide in Table~\ref{tab:classes_names} the  lists of classes contained in each of the target datasets we used for evaluation. 
Overall, \imnet, the randomly sampled set of 100 leaf classes from Imagenet~\cite{deng2009_imagenet}, is more diversified than \cifar, which mostly contains animal classes. \imnet is visually varied between different types of objects, foods, animals, vehicles, clothes and events. \cifar contains, in addition to animals, some types of objects and vehicles.\\

The \food dataset and \birds (extracted from ImageNet~\cite{deng2009_imagenet}) are more specialized than \imnet and \cifar and are thus useful to test the robustness of our method on finer-grained datasets. \places and \food are target datasets which have a larger domain shift with ImageNet classes, and are thus useful to test the robustness of our method against domain variation. Similarly to \imnet, reference datasets are random subsets of ImageNet leaves. They contain various object types and are useful for knowledge transfer.
\vfill

\section{Effects of adaptive bias correction}
\label{sec:state_wise_accs}

In Figure \ref{fig:state_wise_acc}, we illustrate the effects of \adbic on state-wise accuracies, for all backbone IL methods evaluated in this work. Before adaptive correction (\textit{top}), all methods provide strong performance on the last group of classes learned (represented by the diagonals). Their performance is generally poorer for past classes (under the diagonals).

After correction (\textit{bottom}), all methods perform better on past class groups (with a trade-off in accuracy on the last class group) resulting in a higher overall performance.

\section{Effect of the number of reference datasets}
\label{sec:varying_nb_datasets}

In Table \ref{tab:ablation_datasets_2}, we provide a full set of results for the accuracies of \adbic with \lwf, \lucir, \siw and \ftplus when varying the number $R$ of reference datasets, following Table 3 of the main paper. For all methods considered, a single reference dataset is sufficient to obtain significant gains with \adbic.

\begin{table*}[h]
\begin{center}

\begin{subtable}{\linewidth}
    \resizebox{\linewidth}{!}{
        \begin{tabular}{@{\kern0.5em}cccccccccccc@{\kern0.5em}}
            \toprule
            \multirow{2}{*}{\textbf{\textit{S}\ =\ 5}}  & Raw & \textit{R}\ =\ 1 & \textit{R}\ =\ 2 & \textit{R}\ =\ 3 & \textit{R}\ =\ 4 & \textit{R}\ =\ 5 & \textit{R}\ =\ 6 & \textit{R}\ =\ 7 & \textit{R}\ =\ 8 & \textit{R}\ =\ 9 & \textit{R}\ =\ 10 \\ \cmidrule(l){2-12} &
                    
            53.0 & 54.3 \tabnoteb{$\pm$ 0.2} & 54.3 \tabnoteb{$\pm$ 0.2} & 54.3 \tabnoteb{$\pm$ 0.1} & 54.4 \tabnoteb{$\pm$ 0.1} & 54.3 \tabnoteb{$\pm$ 0.1} & 54.3 \tabnoteb{$\pm$ 0.1} & 54.3 \tabnoteb{$\pm$ 0.1} & 54.3 \tabnoteb{$\pm$ 0.1} & 54.3 \tabnoteb{$\pm$ 0.1} & 54.3
            \\ \midrule \midrule

            \multirow{2}{*}{\textbf{\textit{S}\ =\ 10}} & Raw & \textit{R}\ =\ 1 & \textit{R}\ =\ 2 & \textit{R}\ =\ 3 & \textit{R}\ =\ 4 & \textit{R}\ =\ 5 & \textit{R}\ =\ 6 & \textit{R}\ =\ 7 & \textit{R}\ =\ 8 & \textit{R}\ =\ 9 & \textit{R}\ =\ 10 \\ \cmidrule(l){2-12} &
            
            44.0 & 46.2 \tabnoteb{$\pm$ 0.3} & 46.4 \tabnoteb{$\pm$ 0.2} & 46.4 \tabnoteb{$\pm$ 0.2} & 46.4 \tabnoteb{$\pm$ 0.2} & 46.4 \tabnoteb{$\pm$ 0.1} & 46.4 \tabnoteb{$\pm$ 0.1} & 46.5 \tabnoteb{$\pm$ 0.1} & 46.4 \tabnoteb{$\pm$ 0.1} & 46.4 \tabnoteb{$\pm$ 0.1} & 46.4
            \\ \midrule \midrule

            \multirow{2}{*}{\textbf{\textit{S}\ =\ 20}} & Raw & \textit{R}\ =\ 1 & \textit{R}\ =\ 2 &
            \textit{R}\ =\ 3 & \textit{R}\ =\ 4 & \textit{R}\ =\ 5 & \textit{R}\ =\ 6 & \textit{R}\ =\ 7 & \textit{R}\ =\ 8 & \textit{R}\ =\ 9 & \textit{R}\ =\ 10 \\ \cmidrule(l){2-12} &
            
            29.1 & 31.8 \tabnoteb{$\pm$ 0.3} & 32.1 \tabnoteb{$\pm$ 0.1} & 32.1 \tabnoteb{$\pm$ 0.2} & 32.1 \tabnoteb{$\pm$ 0.1} & 32.2 \tabnoteb{$\pm$ 0.1} & 32.2 \tabnoteb{$\pm$ 0.1} & 32.3 \tabnoteb{$\pm$ 0.1} & 32.3 \tabnoteb{$\pm$ 0.1} & 32.3 \tabnoteb{$\pm$ 0.1} & 32.3
            \\ \bottomrule
        \end{tabular}
    }
    \subcaption{\ LwF \cite{li2016_lwf}}
\end{subtable}

\vspace{2em}

\begin{subtable}{\linewidth}
    \resizebox{\linewidth}{!}{
        \begin{tabular}{@{\kern0.5em}cccccccccccc@{\kern0.5em}}
        \toprule
        
        \multirow{2}{*}{\textbf{\textit{S}\ =\ 5}}  & Raw & \textit{R}\ =\ 1 & \textit{R}\ =\ 2 & \textit{R}\ =\ 3 & \textit{R}\ =\ 4 & \textit{R}\ =\ 5 & \textit{R}\ =\ 6 & \textit{R}\ =\ 7 & \textit{R}\ =\ 8 & \textit{R}\ =\ 9 & \textit{R}\ =\ 10 \\ \cmidrule(l){2-12} &
                
        50.1 & 54.7 \tabnoteb{$\pm$ 0.4} & 54.8 \tabnoteb{$\pm$ 0.3} & 54.8 \tabnoteb{$\pm$ 0.1} & 54.8 \tabnoteb{$\pm$ 0.1} & 54.8 \tabnoteb{$\pm$ 0.1} & 54.8 \tabnoteb{$\pm$ 0.1} & 54.8 \tabnoteb{$\pm$ 0.1} & 54.8 \tabnoteb{$\pm$ 0.1} & 54.8 \tabnoteb{$\pm$ 0.1} & 54.8
        \\ \midrule \midrule

        \multirow{2}{*}{\textbf{\textit{S}\ =\ 10}} & Raw & \textit{R}\ =\ 1 & \textit{R}\ =\ 2 & \textit{R}\ =\ 3 & \textit{R}\ =\ 4 & \textit{R}\ =\ 5 & \textit{R}\ =\ 6 & \textit{R}\ =\ 7 & \textit{R}\ =\ 8 & \textit{R}\ =\ 9 & \textit{R}\ =\ 10 \\ \cmidrule(l){2-12} &
        
        33.7 & 42.0 \tabnoteb{$\pm$ 0.7} & 42.1 \tabnoteb{$\pm$ 0.3} & 42.2 \tabnoteb{$\pm$ 0.4} & 42.3 \tabnoteb{$\pm$ 0.3} & 42.2 \tabnoteb{$\pm$ 0.2} & 42.2 \tabnoteb{$\pm$ 0.2} & 42.2 \tabnoteb{$\pm$ 0.1} & 42.2 \tabnoteb{$\pm$ 0.1} & 42.2 \tabnoteb{$\pm$ 0.1} & 42.2
        \\ \midrule \midrule

        \multirow{2}{*}{\textbf{\textit{S}\ =\ 20}} & Raw & \textit{R}\ =\ 1 & \textit{R}\ =\ 2 &
        \textit{R}\ =\ 3 & \textit{R}\ =\ 4 & \textit{R}\ =\ 5 & \textit{R}\ =\ 6 & \textit{R}\ =\ 7 & \textit{R}\ =\ 8 & \textit{R}\ =\ 9 & \textit{R}\ =\ 10 \\ \cmidrule(l){2-12} &
        
        19.5 & 27.5 \tabnoteb{$\pm$ 1.4} & 27.8 \tabnoteb{$\pm$ 0.7} & 27.8 \tabnoteb{$\pm$ 0.9} & 28.3 \tabnoteb{$\pm$ 0.4} & 28.5 \tabnoteb{$\pm$ 0.5} & 28.6 \tabnoteb{$\pm$ 0.6} & 28.5 \tabnoteb{$\pm$ 0.4} & 28.4 \tabnoteb{$\pm$ 0.3} & 28.4 \tabnoteb{$\pm$ 0.2} & 28.4
        \\ \bottomrule
        \end{tabular}
    }
    \subcaption{\ LUCIR \cite{hou2019_lucir}}
\end{subtable}

\vspace{2em}

\begin{subtable}{\linewidth}
    \resizebox{\linewidth}{!}{
        \begin{tabular}{@{\kern0.5em}cccccccccccc@{\kern0.5em}}
            \toprule
            
            \multirow{2}{*}{\textbf{\textit{S}\ =\ 5}}  & Raw & \textit{R}\ =\ 1 & \textit{R}\ =\ 2 & \textit{R}\ =\ 3 & \textit{R}\ =\ 4 & \textit{R}\ =\ 5 & \textit{R}\ =\ 6 & \textit{R}\ =\ 7 & \textit{R}\ =\ 8 & \textit{R}\ =\ 9 & \textit{R}\ =\ 10 \\ \cmidrule(l){2-12} &
                    
            29.9 & 31.6 \tabnoteb{$\pm$ 0.2} & 31.6 \tabnoteb{$\pm$ 0.2} & 31.6 \tabnoteb{$\pm$ 0.1} & 31.7 \tabnoteb{$\pm$ 0.2} & 31.7 \tabnoteb{$\pm$ 0.1} & 31.7 \tabnoteb{$\pm$ 0.1} & 31.7 \tabnoteb{$\pm$ 0.1} & 31.7 \tabnoteb{$\pm$ 0.1} & 31.7 \tabnoteb{$\pm$ 0.1} & 31.7
            \\ \midrule \midrule

            \multirow{2}{*}{\textbf{\textit{S}\ =\ 10}} & Raw & \textit{R}\ =\ 1 & \textit{R}\ =\ 2 & \textit{R}\ =\ 3 & \textit{R}\ =\ 4 & \textit{R}\ =\ 5 & \textit{R}\ =\ 6 & \textit{R}\ =\ 7 & \textit{R}\ =\ 8 & \textit{R}\ =\ 9 & \textit{R}\ =\ 10 \\ \cmidrule(l){2-12} &
            
            22.7 & 23.8 \tabnoteb{$\pm$ 0.4} & 23.8 \tabnoteb{$\pm$ 0.2} & 23.9 \tabnoteb{$\pm$ 0.2} & 24.0 \tabnoteb{$\pm$ 0.2} & 23.9 \tabnoteb{$\pm$ 0.1} & 24.0 \tabnoteb{$\pm$ 0.1} & 24.1 \tabnoteb{$\pm$ 0.1} & 24.0 \tabnoteb{$\pm$ 0.1} & 24.1 \tabnoteb{$\pm$ 0.1} & 24.1
            \\ \midrule \midrule

            \multirow{2}{*}{\textbf{\textit{S}\ =\ 20}} & Raw & \textit{R}\ =\ 1 & \textit{R}\ =\ 2 &
            \textit{R}\ =\ 3 & \textit{R}\ =\ 4 & \textit{R}\ =\ 5 & \textit{R}\ =\ 6 & \textit{R}\ =\ 7 & \textit{R}\ =\ 8 & \textit{R}\ =\ 9 & \textit{R}\ =\ 10 \\ \cmidrule(l){2-12} &
            
            14.8 & 15.7 \tabnoteb{$\pm$ 0.3} & 15.7 \tabnoteb{$\pm$ 0.2} & 15.7 \tabnoteb{$\pm$ 0.2} & 15.8 \tabnoteb{$\pm$ 0.1} & 15.8 \tabnoteb{$\pm$ 0.2} & 15.8 \tabnoteb{$\pm$ 0.1} & 15.8 \tabnoteb{$\pm$ 0.1} & 15.8 \tabnoteb{$\pm$ 0.1} & 15.8 \tabnoteb{$\pm$ 0.1} & 15.8
            \\ \bottomrule
        \end{tabular}
    }
    \subcaption{\ SIW \cite{belouadah2020_siw}}
\end{subtable}

\vspace{2em}

\begin{subtable}{\linewidth}
    \resizebox{\linewidth}{!}{
        \begin{tabular}{@{\kern0.5em}cccccccccccc@{\kern0.5em}}
            \toprule
        
            \multirow{2}{*}{\textbf{\textit{S}\ =\ 5}}  & Raw & \textit{R}\ =\ 1 & \textit{R}\ =\ 2 & \textit{R}\ =\ 3 & \textit{R}\ =\ 4 & \textit{R}\ =\ 5 & \textit{R}\ =\ 6 & \textit{R}\ =\ 7 & \textit{R}\ =\ 8 & \textit{R}\ =\ 9 & \textit{R}\ =\ 10 \\ \cmidrule(l){2-12} &
                    
            28.9 & 31.9 \tabnoteb{$\pm$ 0.2} & 32.0 \tabnoteb{$\pm$ 0.1} & 32.0 \tabnoteb{$\pm$ 0.1} & 32.0 \tabnoteb{$\pm$ 0.1} & 32.0 \tabnoteb{$\pm$ 0.1} & 32.0 \tabnoteb{$\pm$ 0.1} & 31.9 \tabnoteb{$\pm$ 0.1} & 32.0 \tabnoteb{$\pm$ 0.1} & 32.0 \tabnoteb{$\pm$ 0.1} & 31.9
            \\ \midrule \midrule

            \multirow{2}{*}{\textbf{\textit{S}\ =\ 10}} & Raw & \textit{R}\ =\ 1 & \textit{R}\ =\ 2 & \textit{R}\ =\ 3 & \textit{R}\ =\ 4 & \textit{R}\ =\ 5 & \textit{R}\ =\ 6 & \textit{R}\ =\ 7 & \textit{R}\ =\ 8 & \textit{R}\ =\ 9 & \textit{R}\ =\ 10 \\ \cmidrule(l){2-12} &
            
            22.6 & 23.2 \tabnoteb{$\pm$ 0.4} & 23.5 \tabnoteb{$\pm$ 0.2} & 23.5 \tabnoteb{$\pm$ 0.2} & 23.6 \tabnoteb{$\pm$ 0.1} & 23.5 \tabnoteb{$\pm$ 0.2} & 23.6 \tabnoteb{$\pm$ 0.1} & 23.6 \tabnoteb{$\pm$ 0.1} & 23.6 \tabnoteb{$\pm$ 0.1} & 23.6 \tabnoteb{$\pm$ 0.1} & 23.6
            \\ \midrule \midrule

            \multirow{2}{*}{\textbf{\textit{S}\ =\ 20}} & Raw & \textit{R}\ =\ 1 & \textit{R}\ =\ 2 &
            \textit{R}\ =\ 3 & \textit{R}\ =\ 4 & \textit{R}\ =\ 5 & \textit{R}\ =\ 6 & \textit{R}\ =\ 7 & \textit{R}\ =\ 8 & \textit{R}\ =\ 9 & \textit{R}\ =\ 10 \\ \cmidrule(l){2-12} &
            
            14.5 & 14.8 \tabnoteb{$\pm$ 0.2} & 15.0 \tabnoteb{$\pm$ 0.1} & 15.0 \tabnoteb{$\pm$ 0.2} & 15.1 \tabnoteb{$\pm$ 0.1} & 15.0 \tabnoteb{$\pm$ 0.1} & 15.1 \tabnoteb{$\pm$ 0.1} & 15.1 \tabnoteb{$\pm$ 0.1} & 15.0 \tabnoteb{$\pm$ 0.1} & 15.0 \tabnoteb{$\pm$ 0.1} & 15.0
            \\ \bottomrule
        \end{tabular}
    }
    \subcaption{\ FT+ \cite{masana2021_study}}
\end{subtable}

\vspace{1.5em}

\caption{Average top-1 incremental accuracy of \adbic-corrected models trained incrementally on \cifar with \lwf, \lucir, \siw and \ftplus, for $S=\{5, 10, 20\}$ states, while varying the number $R$ of reference datasets. For $R \leq 9$, results are averaged across 10 random samplings of the reference datasets. \textit{Raw} is the accuracy of each method without bias correction.}
\label{tab:ablation_datasets_2}

\end{center}
\end{table*}

\end{document}